\documentclass[journal]{IEEEtran}
\usepackage[bookmarks=true]{hyperref}
\usepackage{hyperref}       
\usepackage{url}            
\usepackage{booktabs}       
\usepackage{amsfonts}       
\usepackage{nicefrac}       
\usepackage{microtype}      
\usepackage{algorithm}
\usepackage{algorithmic}
\usepackage{graphicx}
\usepackage{makecell}
\usepackage{subfigure}
\usepackage{multirow}
\usepackage{multicol}
\usepackage{bm}
\usepackage{booktabs} 
\usepackage{pdfpages}
\usepackage{overpic}
\usepackage{pict2e}
\usepackage{subfigure}
\usepackage{amsmath}
\usepackage{xcolor}
\definecolor{mygray}{gray}{0.45}
\usepackage{textcomp}

\usepackage{amssymb}

\usepackage{pifont}

\RequirePackage{silence}
\graphicspath{{figures/}}

\ifdefined \GramaCheck
\newcommand{\CheckRmv}[1]{}
\newcommand{\figref}[1]{Figure 1}%
\newcommand{\tabref}[1]{Table 1}%
\newcommand{\secref}[1]{Section 1}
\renewcommand{\eqref}[1]{Equation 1}
\else
\newcommand{\CheckRmv}[1]{#1}
\newcommand{\figref}[1]{Fig.~\ref{#1}}%
\newcommand{\tabref}[1]{Tab.~\ref{#1}}%
\newcommand{\secref}[1]{Sec.~\ref{#1}}
\renewcommand{\eqref}[1]{Eqn.~(\ref{#1})}
\fi

\def\ie{\emph{i.e.,~}}

\def\etal{{\em et al.}}

\begin{document}
\title{Delving Deep into Label Smoothing}
\author{Chang-Bin~Zhang$^\dagger$,
	Peng-Tao~Jiang$^\dagger$,
	Qibin~Hou,
	Yunchao~Wei,
	Qi~Han,
	Zhen~Li,
	and~Ming-Ming~Cheng
	\IEEEcompsocitemizethanks{
		\IEEEcompsocthanksitem C.B.~Zhang, P.T.~Jiang, Q.~Han, Z.~Li and M.M.~Cheng
		are with TKLNDST, CS, Nankai University.
		M.M.~Cheng is the corresponding author (cmm@nankai.edu.cn).
		$\dagger$ denotes equal contribution. 
		\IEEEcompsocthanksitem Q.~Hou is with the National University of Singapore.
		\IEEEcompsocthanksitem Y.~Wei is with the Beijing Jiaotong University.
	}
}

\markboth{IEEE Transactions on Image Processing,~2021,~Vol.~30,~Pages 5984-5996,~DOI~10.1109/TIP.2021.3089942}%
{Zhang \MakeLowercase{\textit{et al.}}: Bare Demo of IEEEtran.cls for IEEE Journals}
%

\maketitle

\begin{abstract}
Label smoothing is an effective regularization tool for 
deep neural networks (DNNs), which generates soft labels 
by applying a weighted average between the uniform distribution
and the hard label.
It is often used to reduce the overfitting problem of 
training DNNs and further improve classification performance.
In this paper, we aim to investigate how to generate 
more reliable soft labels.
We present an Online Label Smoothing (OLS) strategy, 
which generates soft labels based on the statistics of 
the model prediction for the target category.
The proposed OLS constructs a more reasonable probability 
distribution between the target categories and non-target 
categories to supervise DNNs.
Experiments demonstrate that based on the same classification 
models, the proposed approach can effectively 
improve the classification performance on CIFAR-100, ImageNet, 
and fine-grained datasets. 
Additionally, the proposed method can significantly improve 
the robustness of DNN models to noisy labels compared to current 
label smoothing approaches.
The source code is available at our project page:
\url{https://mmcheng.net/ols/} and \url{https://github.com/zhangchbin/OnlineLabelSmoothing}
\end{abstract}



\section{Introduction}\label{sec:introduction}
\IEEEPARstart{D}{eep} Neural Networks (DNNs)~\cite{vgg,he2016deep,huang2017densely,xie2017aggregated,hu2018squeeze,sandler18mobile,gao2019res2net} 
have achieved remarkable performance in image recognition
~\cite{krizhevsky2009learning,deng2009imagenet}. 
However, most DNNs tend to fall into over-confidence for training samples, 
greatly influencing their generalization ability to test samples.
Recently, researchers have proposed many regularization
approaches,
including Label Smoothing~\cite{christian2016rethinking},
Bootstrap~\cite{bootstrap},
CutOut~\cite{devries2017cutout},
MixUp~\cite{zhang2018mixup},
DropBlock~\cite{ghiasi2018dropblock} and
ShakeDrop~\cite{yamada2019shakedrop},
to conquer the overfitting problem to the distribution of 
the training set.
These methods attempt to tackle this problem from the views of data augmentation \cite{devries2017cutout,zhang2018mixup}, 
model design \cite{ghiasi2018dropblock,yamada2019shakedrop},
or label transformation \cite{christian2016rethinking,bootstrap,qi2020loss}.
Among them, label smoothing is a simple yet effective
regularization tool operating on the labels.

Label smoothing (LS), 
aiming at providing regularization for a learnable classification model, 
is first proposed in~\cite{christian2016rethinking}.
Instead of merely leveraging the hard labels 
for training (\figref{f1a}), 
Christian \etal \cite{christian2016rethinking} utilizes soft labels 
by taking an average between the 
hard labels and the uniform distribution over labels (\figref{f1b}).
Although such kind of soft labels can provide strong regularization and prevent the learned models from being 
over-confident, it treats the non-target categories equally 
by assigning them with fixed identical probability.
For example, a `cat' should be more like a `dog' rather than 
an `automobile.'
Therefore, we argue that the assigned probabilities of
non-target categories should highly consider their 
similarities to the category of the given image.
Equally treating each non-target category could weaken
the capability of label smoothing and limit the model performance.

\newcommand{\legnd}[2]{#1 -- \textsl{#2} ~~~~~} 
\newcommand{\addImg}[2]{\label{#2}\includegraphics[width=0.31\linewidth]{labels/#1}}

\begin{figure*}[!th] 
	\centering
	\label{f1}
	\subfigure[Hard Label]{\addImg{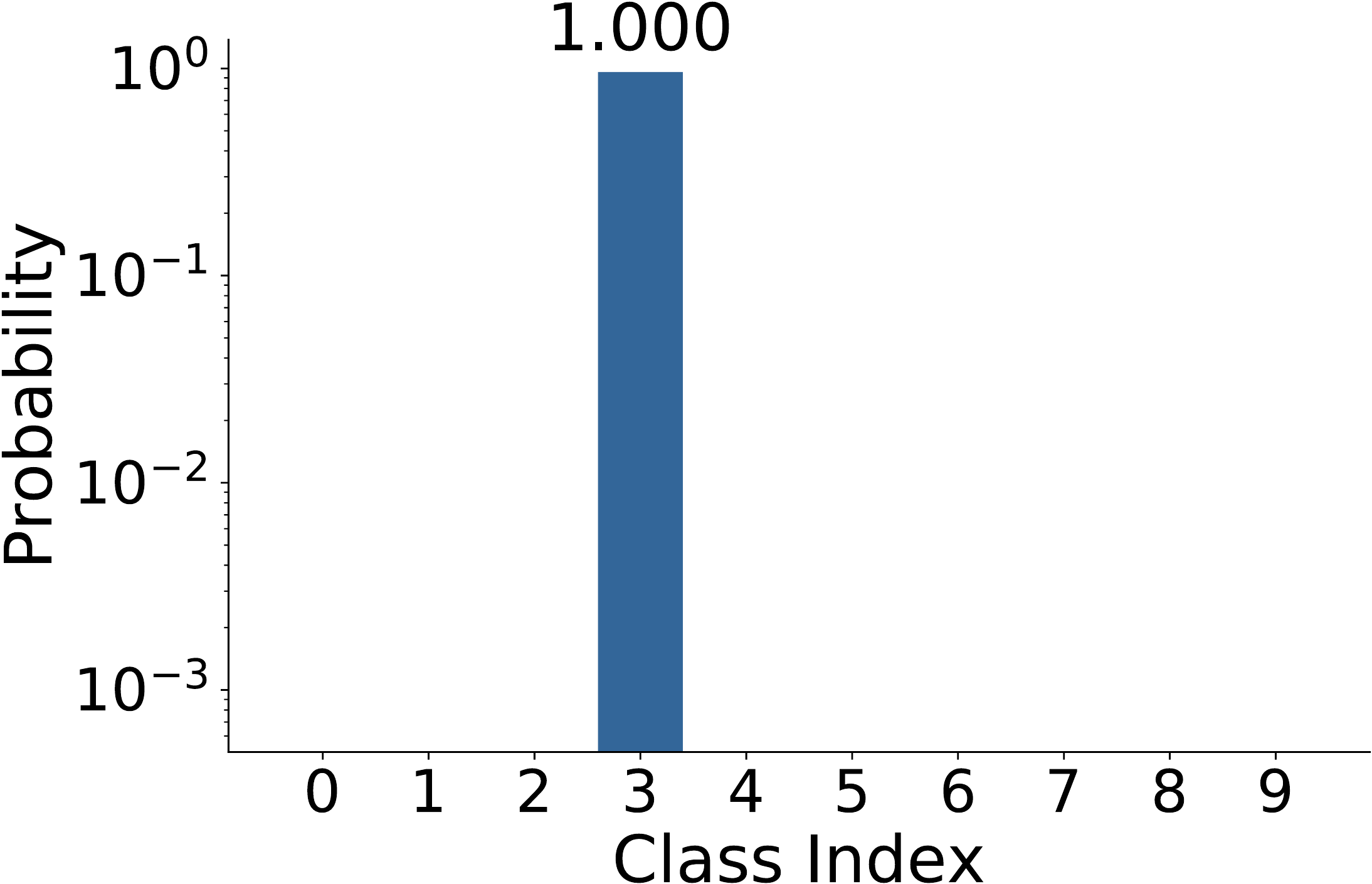}{f1a}}
	\subfigure[LS]{\addImg{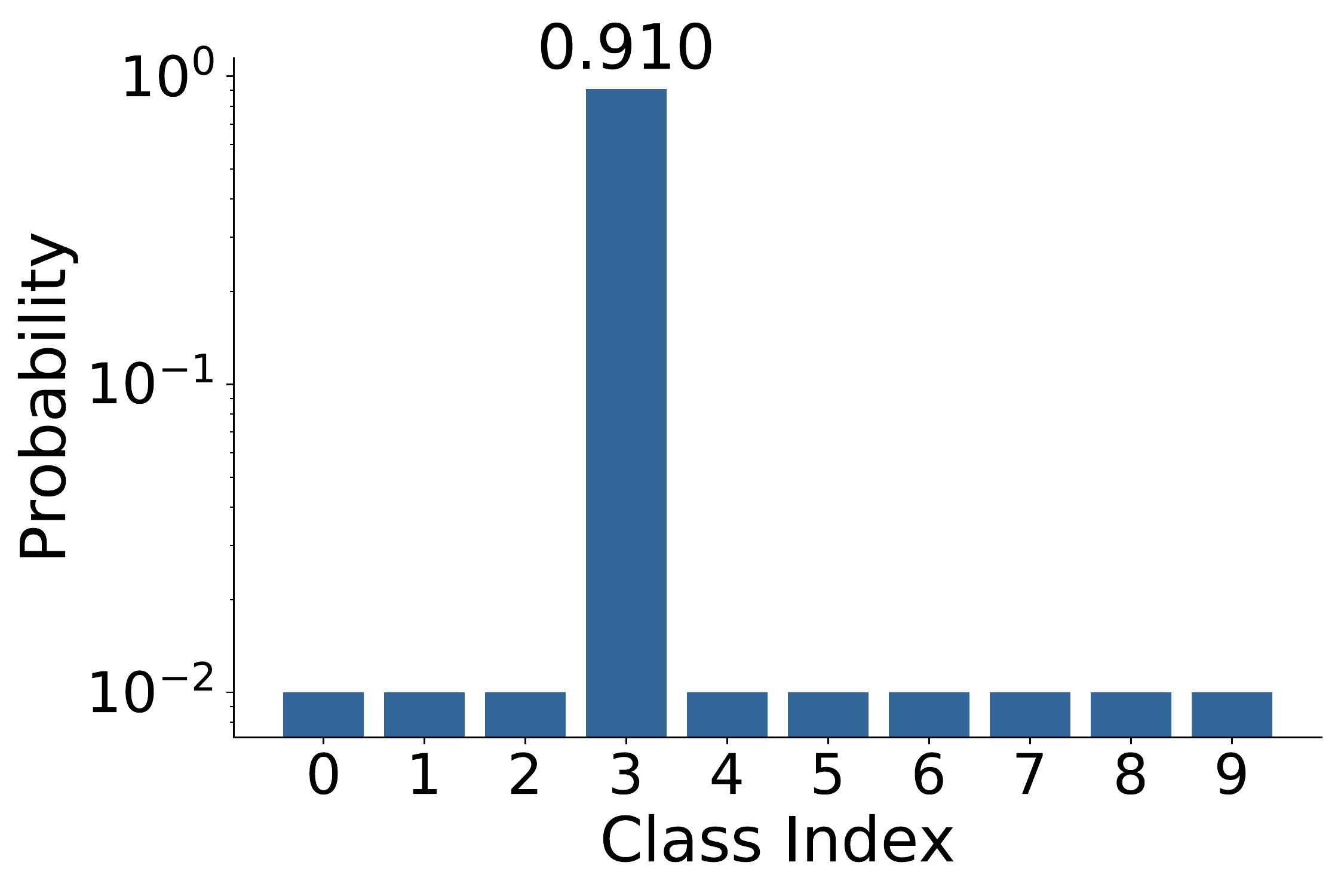}{f1b}}
	\subfigure[Ours]{\addImg{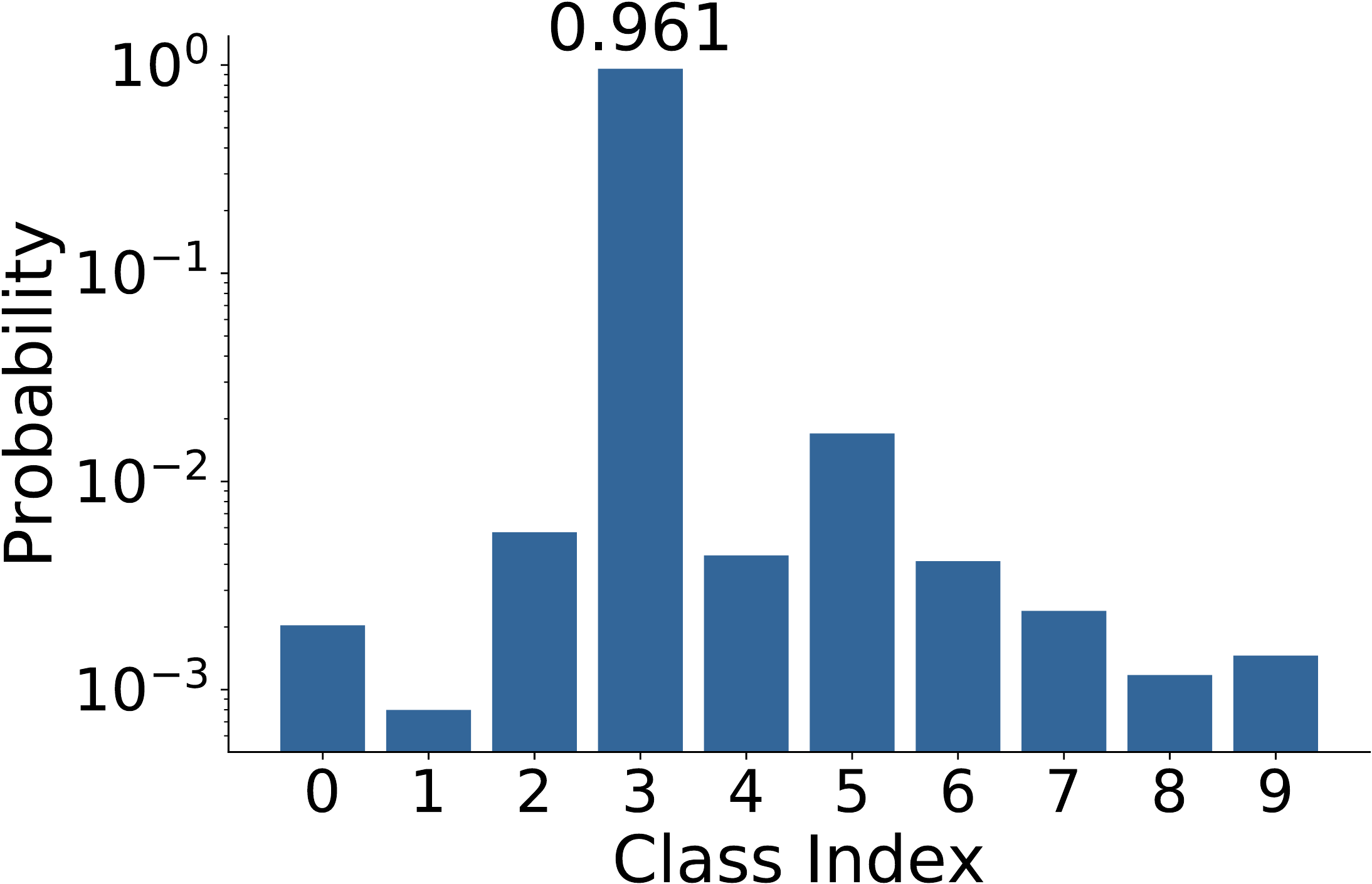}{f1c}}
	\fbox{\scriptsize
		\legnd{0}{airplane} \legnd{1}{automobile}	\legnd{2}{bird}
		\legnd{3}{cat} \legnd{4}{deer} \legnd{5}{dog} \legnd{6}{frog}
		\legnd{7}{horse} \legnd{8}{ship} 9 -- \textsl{truck}
	}
	\caption{Different kinds of label distributions
		on the CIFAR-10 dataset. The target category is `cat.'
		We scale the $y$-axis using the $log$ function 
		for visualization. (a) Original hard label.
		(b) Soft label generated by LS
		\cite{christian2016rethinking}. This soft label is a 
		mixture of the hard label and uniform distribution.
		(c) Soft label generated by our OLS method during 
		the training process of ResNet-29. 
	}
	\vskip -0.1in
\end{figure*}

It has been demonstrated in~\cite{hinton15KD} that 
model predictive distributions provide a promising way 
to reveal the implicit relationships among different
categories.
Motivated by this knowledge, we propose a simple 
yet effective method to generate more reliable soft 
labels that consider the relationships among 
different categories to take the place of label smoothing.
Specifically, we maintain a moving label distribution 
for each category, which can be updated during the training 
process.
The maintained label distributions keep changing 
at each training epoch and are utilized to supervise DNNs 
until the model reaches convergence.
Our method takes advantage of the statistics of the
intermediate model predictions, which can better 
build the relationships between the target categories 
and the non-target ones.
It can be observed from \figref{f1c} that our method
gives more confidence to the animal categories
instead of those non-animal ones when the label is `cat.'

We conduct extensive experiments on CIFAR-100, ImageNet~\cite{deng2009imagenet} and four fine-grained datasets~\cite{nilsback2008automated,WahCUB_200_2011,krause20133d,maji2013fine}.
Our OLS can make consistent improvements over baselines. 
To be specific, directly applying OLS to ResNet-56 and ResNeXt29-2x64d 
yields 1.57\% and 2.11\% top-1 performance gains on CIFAR-100, respectively. 
For ImageNet, our OLS can bring 1.4\% and 1.02\% performance improvements 
to ResNet-50 and ResNet-101 \cite{he2016deep}, respectively. 
On four fine-grained datasets,
OLS achieves an average 1.0\% performance improvement over LS~\cite{christian2016rethinking} on four different backbones, i.e.,
ResNet-50~\cite{he2016deep},
MobileNetv2~\cite{sandler18mobile},
EfficientNet-b7~\cite{efficientNet}
and SAN-15~\cite{zhao2020exploring}.
The proposed OLS can be naturally employed to tackle noisy labels 
by reducing the overfitting to training sets. 
Additionally,
OLS can be conveniently used in the training process of many models.
We hope it can serve as an effective regularization tool 
to augment the training of classification models.

\section{Related Work} \label{sec:related work}

\begin{table*}[!thp]
	\caption{Comparison between our method and knowledge distillation. Self-KD denotes self-Knowledge Distillation.} \label{tab:comKD}
	\begin{center}
		\begin{tabular}{c|cccc}
			\toprule
			~ & Vanilla KD~\cite{hinton15KD}   & Arch.-based self-KD~\cite{zhang2019byot} & Data-based self-KD~\cite{xu2019data} & OLS (Ours) \\
			\midrule
			Label  & sample-level & sample-level & sample-level & class-level \\
			\midrule
			Trained teacher model                   & \ding{51} & \ding{55}  & \ding{55} & \ding{55}\\
			\midrule
			Special network architecture            & \ding{55} & \ding{51}  & \ding{55} & \ding{55} \\ 
			\midrule
			Forward times in one training iteration & 2 & 1  & 2  & 1 \\
			\bottomrule
		\end{tabular}
	\end{center}
	\vskip -0.2in
\end{table*}

\textbf{Regularization tools on labels.}
Training DNNs with hard labels (assigning 1 
to the target category and 0 to the non-target ones)
often results in over-confident models. 
Boosting labels is a straightforward yet effective way 
to alleviate the overfitting problem and improve the accuracy 
and robustness of DNNs.
Bootstrapping~\cite{bootstrap} provided two options, Bootsoft and Boothard, 
which smoothed the hard labels using
the predicted distribution and the predicted class, respectively.
Xie \etal~\cite{Xie2016disturb} randomly perturbed
labels of some samples in a mini-batch to regularize 
the networks.
To further prevent the training models from 
overfitting to some specific samples, 
Dubey \etal~\cite{dubey2018pairwise} added pairwise confusion to the output 
logits of samples belonging to different categories 
in training so that the models can learn slightly less
discriminative features for specific samples.
Li \etal~\cite{li2020reconstruction} used two networks to
embed the images and the labels in a latent space
and regularize the network via the distance between these embeddings.
Christian \etal~\cite{christian2016rethinking} leveraged soft labels 
for training, where the soft labels are generated 
by taking an average between the hard labels 
and the uniform distribution over labels.
Our OLS also focuses on generating soft labels 
that can provide stable regularization for models.
Following AET~\cite{zhang2019aet,qi2020learning} and AVT~\cite{qi2019avt},
Wang~\etal~\cite{wang2020enaet} proposed an innovative framework, EnAET~\cite{wang2020enaet},
that combined semi-supervised and self-supervised training.
It learned feature representation by predicting non-spatial and spatial transformation parameters.
Both our method and EnAET~\cite{wang2020enaet} obtained soft labels by accumulating predictions of multiple samples.
However,
our method is very different from EnAET~\cite{wang2020enaet}.
It obtained soft labels by accumulating the augmented views of the same sample by different transformation functions.
This consistency constraint is also often used in self-Knowledge Distillation.
In contrast,
our method is to encourage the predictions of all samples in the same class to become consistent
by the accumulated class-level soft labels.
Unlike the mentioned approaches above, the soft 
labels generated by OLS take advantage of the statistical 
characteristics of model predictions of intermediate states.

\textbf{Knowledge distillation.}
Knowledge distillation~\cite{hinton15KD,passalis19unsupervised,tommaso2018ban}, 
is a popular way to compress models, which can significantly 
improve the performance of light-weight networks.		
Knowledge distillation has been widely used in
many tasks~\cite{ge2020distilling,wang2020real,ge2019low,peng2019few}.
Hinton~\etal~\cite{hinton15KD} show that the success of knowledge distillation
is due to the model's response to the non-target classes.
It shows that DNNs can discover the similarities
among different categories~\cite{hinton15KD,tommaso2018ban}
hidden in the predictions.
Inspired by knowledge distillation, some works 
\cite{zhang2019byot,xu2019data,tommaso2018ban}
utilized a self-distillation strategy to improve 
classification accuracy. 
BYOT~\cite{zhang2019byot} designed a network architecture-based self-Knowledge Distillation,
which distilled the knowledge from the deep layers to the shallow layer.
Xu \etal~\cite{xu2019data} applied a data-based self-Knowledge Distillation
and encouraged the output of the augmented samples (using data augmentation methods)
to be consistent with the original samples.
Furlanello \etal~\cite{tommaso2018ban} proposed to
distill the knowledge of the teacher model
to the student model with the same architecture.
The student model obtained a higher accuracy 
than the teacher model.
At the same time, Tommaso \etal~\cite{tommaso2018ban} also 
verified the importance of the similarity between categories 
in the soft labels.
Our work is inspired by knowledge distillation,
aiming to find a reasonable similarity among categories.
Both knowledge distillation and our method use the output logits of the network
as soft labels and benefit from the similarities hidden in the logits~\cite{hinton15KD,tommaso2018ban}.
But there are many differences between our method and the knowledge distillation.
We summarize the main differences in~\tabref{tab:comKD}.
Without any teacher models,
compared with knowledge distillation,
our method could save the training cost,
\ie{our method does not bring extra forward propagations}.
Besides,
our method is applicable to any network architecture without special modification.

\textbf{Classification against noisy labels.}
Noisy labels in current datasets are inevitable due to
the incorrect annotations by humans. 
To deal with this problem, many researchers
explored solutions to this problem
from both models~\cite{yao2019deep,duncan1992reinforcement},
data~\cite{wang2018multiclass,wei2020harness}
and training strategies~\cite{han2018progressive,tanaka2018joint,han2019deep}.
A typical idea~\cite{Ren2018reweight,metaweight,liu2015classification} is to 
weight different samples to reduce the influence of noisy samples 
on training.
Ren \etal~\cite{Ren2018reweight} verified each mini-batch on 
the clean validation set to adjust each sample's weight 
in a mini-batch dynamically.
MetaWeightNet~\cite{metaweight} also exploited the clean
validation set to learn the weights for samples by
a multilayer perceptron (MLP).
Moreover, some researchers solve this problem from the optimization 
perspective~\cite{sl,tanno2019learn}.
Wang \etal~\cite{sl} improved the 
robustness against noisy labels by replacing the normal 
cross-entropy function with the symmetric cross-entropy function.
Arazo \etal~\cite{unsupervised} observed that noisy
samples often have higher losses than 
the clean ones during the early epochs of training.
Based on this observation, they proposed to use the beta mixture
model to represent clean samples and noisy samples
and adopt 
this model to provide estimates of the actual class 
for noisy samples.
Another kind of idea \cite{zhang2018improving,fang2018data}
is to train the network with only the right labels.
PENCIL~\cite{yi2019probabilistic} proposed a novel framework
to learn the correct label and model's weights at the same time.
This method maintained a learnable label for each sample.
Han~\etal~\cite{han2019deep} designed the label correction phase and
performed the training phase and label correction phase iteratively.
They got multiple prototypes for each class and redefined the labels for all samples.
Different from these two methods,
our method does not specifically design the process of label correction.
Therefore,
our method does not bring extra learnable parameters and
does not conflict with the label correction strategy designed in Han~\etal~\cite{han2019deep}.
On the other hand,
we accumulate the output of correctly predicted samples during training to get the soft labels for each class.
These soft labels bring intra-class constraints to reduce the over-fitting to the wrong labels,
which improves the robustness to noisy labels.
Although the proposed OLS is not specifically designed 
for noisy labels, the classification accuracy on noisy 
datasets is largely improved when training models with OLS.
The performance gain owes to the ability of OLS 
to reduce the overfitting to noisy samples.

\begin{figure*}[!th] 
	\begin{small} \label{f2}
		\begin{overpic}[width=1.0\textwidth, tics=2]{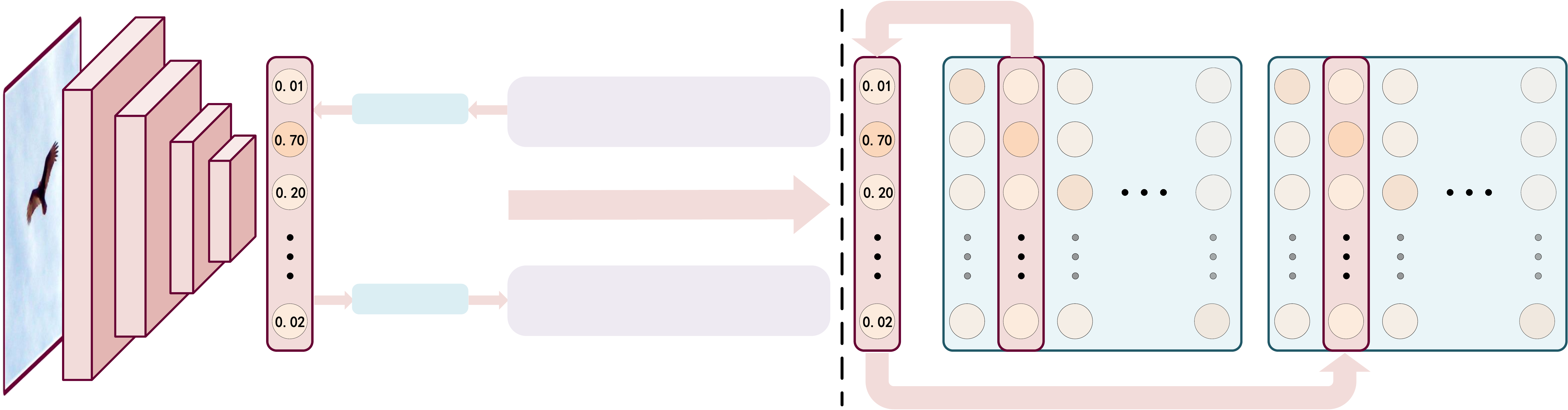}
			\put(22.5, 18.7){\textbf{Supervise}}
			\put(23.3, 6.7){\textbf{Update}}
			
			\textcolor{mygray}{\put(32.4, 6.5) {$S^1$}}
			\put(35.4, 6.5) {$S^2$}
			\textcolor{mygray}{
				\put(38.4, 6.5) {$S^3$}	
				\put(41.4, 6.5) {$\cdot$}
				\put(41.8, 6.5) {$\cdot$}	
				\put(42.2, 6.5) {$\cdot$}
				\put(43.6, 6.5) {$S^{t}$}	
				\put(46.4, 6.5) {$\cdot$}
				\put(46.8, 6.5) {$\cdot$}
				\put(47.2, 6.5) {$\cdot$}
				\put(48.8, 6.5) {$S^{T}$}
			}
			
			\textcolor{mygray}{
				\put(32, 18.5) {$S^0$}
			}
			\put(35, 18.5) {$S^1$}
			\textcolor{mygray}{
				\put(38, 18.5) {$S^2$}
				\put(41, 18.5) {$\cdot$}
				\put(41.4, 18.5) {$\cdot$}
				\put(41.8, 18.5) {$\cdot$}
				\put(42.5, 18.5) {$S^{t-1}$}
				\put(46.2, 18.5) {$\cdot$}
				\put(46.6, 18.5){$\cdot$}
				\put(47, 18.5) {$\cdot$}
				\put(48, 18.5) {$S^{T-1}$}
			} 
			
			\put(26, 12.6) {\textbf{epoch}}
			\textcolor{mygray}{
				\put(32, 12.6) {$\#$1}
			}
			\put(35, 12.6) {$\#$\textbf{2}}
			\textcolor{mygray}{
				\put(38, 12.6) {$\#3$}
				\put(41, 12.6){$\cdot$}
				\put(41.4, 12.6){$\cdot$}
				\put(41.8, 12.6){$\cdot$}
				\put(43, 12.6) {$\#t$}
				\put(46.2, 12.6) {$\cdot$}
				\put(46.6, 12.6){$\cdot$}
				\put(47.0, 12.6){$\cdot$}
				\put(47.6, 12.67){$\#T$}
			}	
			\put(56.4, 24.7) {\textbf{Supervise}}
			\put(67, 0.3) {\textbf{Update}}
			
			\put(68.7, 24) {$S^1$}
			\put(88.7, 24) {$S^2$}
			
			\put(14, 2) {Predicted category: 1}
			\put(14, 0) {Ground-Truth: 1}
			
			\put(60.8, 2){0}
			\put(64.2, 2){1}
			\put(67.6, 2){2}
			\put(71, 2){$\cdots$}
			\put(76.0, 2){$K$}
			
			\put(81.5, 2){0}
			\put(85, 2){1}
			\put(88.5, 2){2}
			\put(91.7, 2){$\cdots$}
			\put(96.7, 2){$K$}
			
		\end{overpic}
	\end{small}
	\vskip -0.1in
	\caption{The illustration of training DNN with our online label smoothing method. 
		The left part of the figure shows the whole training process.
		We simply divided the training process into T phases according to the training epochs. 
		$K$ denotes the number of categories in datasets. 
		We define each column of $S^t$ to represent the soft label for a target category. 
		At each epoch, we use the soft labels generated in the previous epoch to supervise the model, 
		and meanwhile, we generate the soft labels for the next epoch. 
		In the right, we show a detailed example of the training process in epoch$\#2$.
		The generation of $S^t$ is depicted in \secref{sec:methods}.}
\end{figure*}
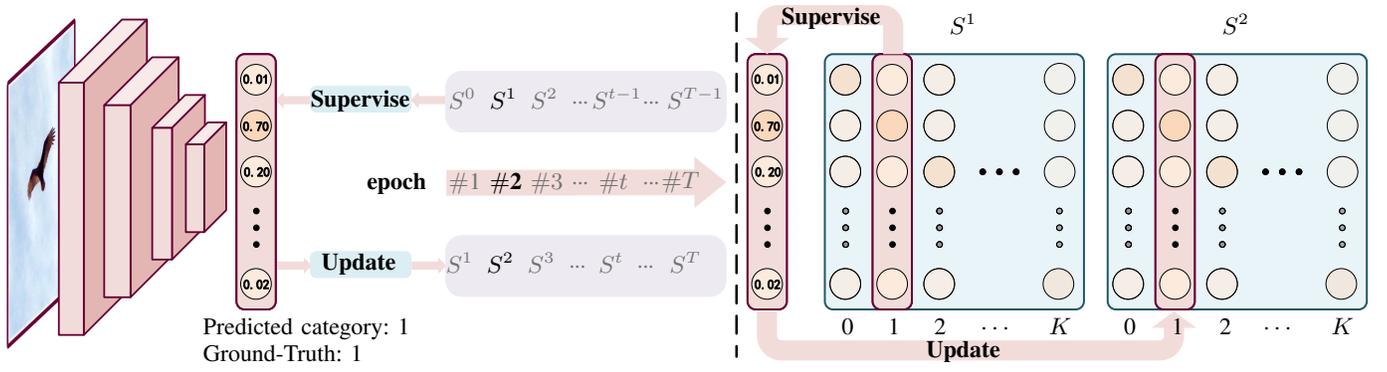
\section{Method} \label{sec:methods}

\subsection{Preliminaries}

Given a dataset $\mathcal{D}_{\tt train}=\{(\bm{x_i}, y_i)\}$
with $K$ classes, where $\bm{x_i}$ denotes 
the input image and $y_i$ denotes the corresponding 
ground-truth label.
For each sample $(\bm{x_i}, y_i)$, the DNN model
predicts a probability $p(k|\bm{x_i})$ for the 
class $k$ using the softmax function.
The distribution $q$ of the hard label $y_i$ can 
be denoted as
$q(k=y_i|\bm{x_i})=1$ and $q(k \ne y_i|\bm{x_i})=0$.
Then, the standard cross-entropy loss used in image
classification for $(\bm{x_i}, y_i)$ can be written as
\begin{equation}
\begin{aligned}
	L_{hard} = - \sum_{k=1}^K q(k|\bm{x_i})\log{p(k|\bm{x_i})} \\
	=-\log{p(k=y_i|\bm{x_i})}. \label{eq:hardce}
\end{aligned}
\end{equation}

Instead of using hard labels for model training,
LS~\cite{christian2016rethinking} utilizes soft labels that are 
generated by exploiting a uniform distribution to smooth 
the distribution of the hard labels.
Specifically, the probability of $\bm{x_i}$ being class $k$ 
in the soft label can be expressed as
\begin{equation}
	q'(k|\bm{x_i})=(1-\varepsilon)q(k|\bm{x_i})+\frac{\varepsilon}{K},
\end{equation}
where $\varepsilon$ denotes the smoothing parameter that \
is usually set to 0.1 in practice.
The assumption behind LS is that
the confidence for the non-target categories is treated 
equally as shown in \figref{f1b}.
Although combining the uniform distribution with the original
hard label is useful for regularization, LS itself
does not consider the genuine relationships among different 
categories~\cite{MullerKH19}.
We take this into account and present our online 
label smoothing method accordingly.

\subsection{Online Label Smoothing} \label{sec:algorithm}

According to knowledge distillation, 
the similarity among categories can be effectively discovered
from the model predictions~\cite{tommaso2018ban,hinton15KD}.
Motivated by this fact, unlike LS utilizing a static soft label, 
we propose to exploit model predictions to continuously update the 
soft labels during the training phase.
Specifically, in the training process, we maintain 
a class-level soft label for each category.
Given an input image $\bm{x_i}$, if the classification is correct, 
the soft label corresponding to the target class $y_i$ will be 
updated using the predicted probability $p(\bm{x_i})$.
Then the updated soft labels will be subsequently utilized
to supervise the model.
The pipeline of our proposed method is shown in \figref{f2}.

Formally, let $T$ denote the number of training epochs.
We then define $\mathcal{S} = \{S^0, S^1, \cdots, S^t, \cdots, S^{T-1}\}$ 
as the collection of the class-level soft labels 
at different training epochs.
Here, $S^t$ is a matrix with $K$ rows and $K$ columns, 
and each column in $S^t$ corresponds to the soft 
label for one category.
In the $t_{th}$ training epoch, given a sample $(\bm{x_i}, y_i)$, 
we use the soft label $S^{t-1}_{y_i}$ to form a temporary label 
distribution to supervise the model, where $S^{t-1}_{y_i}$ denotes 
the soft label for the target category $y_i$.
The training loss of the model supervised by $S^{t-1}_{y_i}$
for $(\bm{x_i}, y_i)$ can be represented by
\begin{equation}
	L_{soft} = - \sum_{k=1}^{K}S^{t-1}_{y_i, k} \cdot \log{p(k|\bm{x_i})}. \label{eq:softce}
\end{equation}

It is possible that we directly use the above soft label to 
supervise the training model, but we find that the model is 
hard to converge due to the random parameter initialization
at the beginning and the lack of the hard label.
Thus, we utilize both the hard 
label and soft label as supervision to train the model. 
Now, the total training loss can be represented by
\begin{equation}
	L = \alpha L_{hard} + (1-\alpha) L_{soft}, \label{eq:olsloss}
\end{equation}
where $\alpha$ is used to balancing $L_{hard}$ and $L_{soft}$.

\begin{algorithm}[t!]
	\caption{\small{The pipeline of the proposed OLS}}
	\label{alg:ols}
	\begin{algorithmic}
		\STATE {\bfseries Input:} Dataset $\mathcal{D}_{\tt train}=\{(\bm{x_i}, y_i)\}$, 
		model $f_{\theta}$, training epochs $T$
		\STATE {\bfseries Initialize:} Soft label matrix $\bm{S^0} = \frac{1}{K}\bm{I}$, 
		$\bm{I}$ denotes unit matrix, $K$ denotes the number of classes
		\FOR{current epoch $t=1$ {\bfseries to} $T$}
		\STATE {\bfseries Initialize:} $\bm{S^t} = \bm{0}$
		
		\FOR{$iter=1$ {\bfseries to} $iterations$}
		\STATE Sample a batch $\mathcal{B}\subset\mathcal{D}_{\tt train}$, input to $f_{\theta}$
		\STATE Obtain predicted probabilities $\{f(\theta,\bm{x_i}), \bm{x_i} \in \mathcal{B}\}$
		\STATE Compute loss by \eqref{eq:olsloss}, backward to update the parameter $\theta$
		\FOR{$i=1$ {\bfseries to} $|\mathcal{B}|$}
		\STATE Update $\bm{S_{y_i}^{t}} \leftarrow \bm{S_{y_i}^{t}} + f(\theta,\bm{x_i})$
		\ENDFOR
		\ENDFOR
		\STATE Normalize $\bm{S^t}$ at each column 
		\ENDFOR
	\end{algorithmic}
\end{algorithm}

\begin{figure*}[!ht]
	\begin{small} 
		\centering
		\begin{minipage}{\linewidth}
			\centering	
			\subfigure[Hard Label]{
				\begin{minipage}{0.31\linewidth}
					\includegraphics[width=\linewidth]{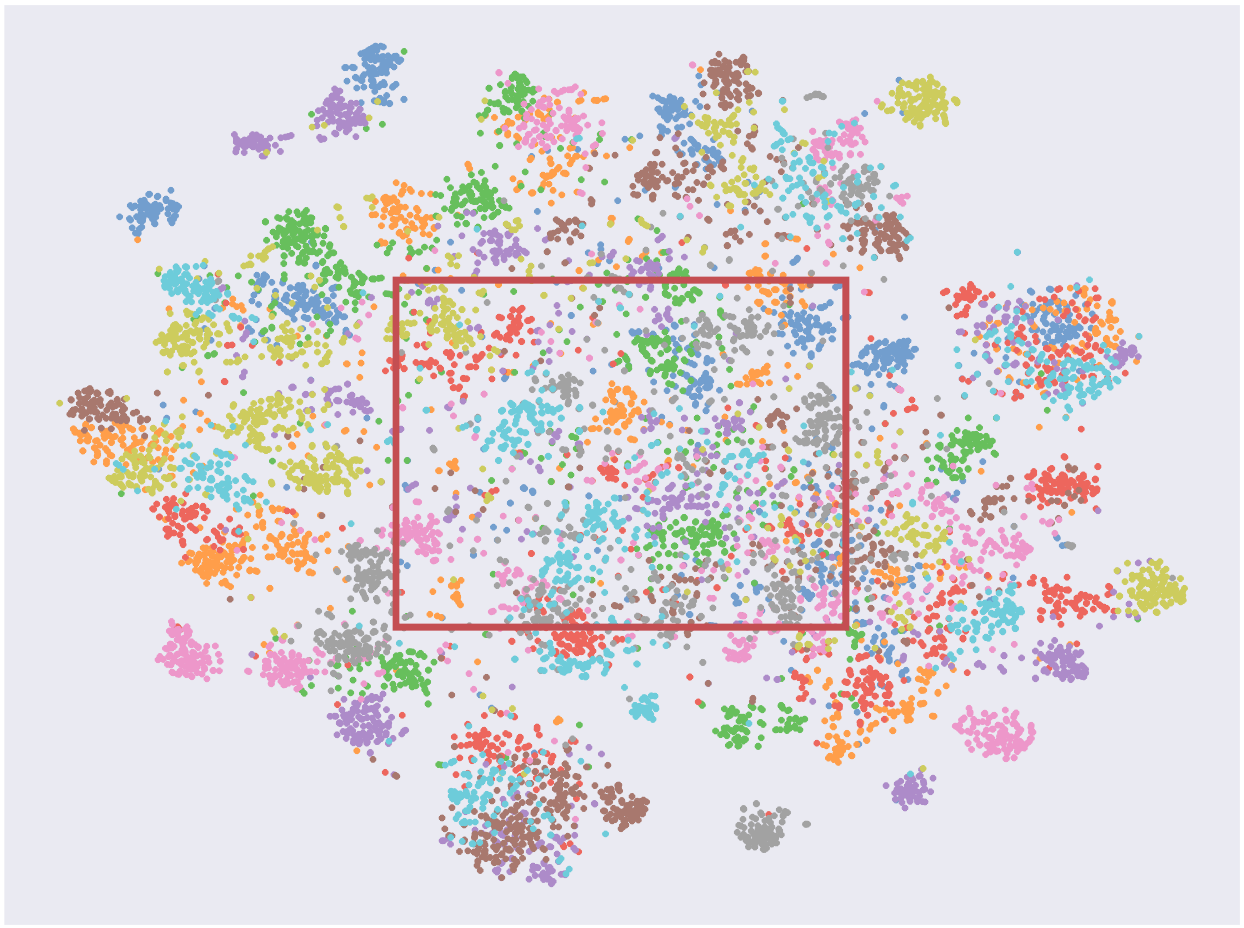}\\
					\includegraphics[width=\linewidth]{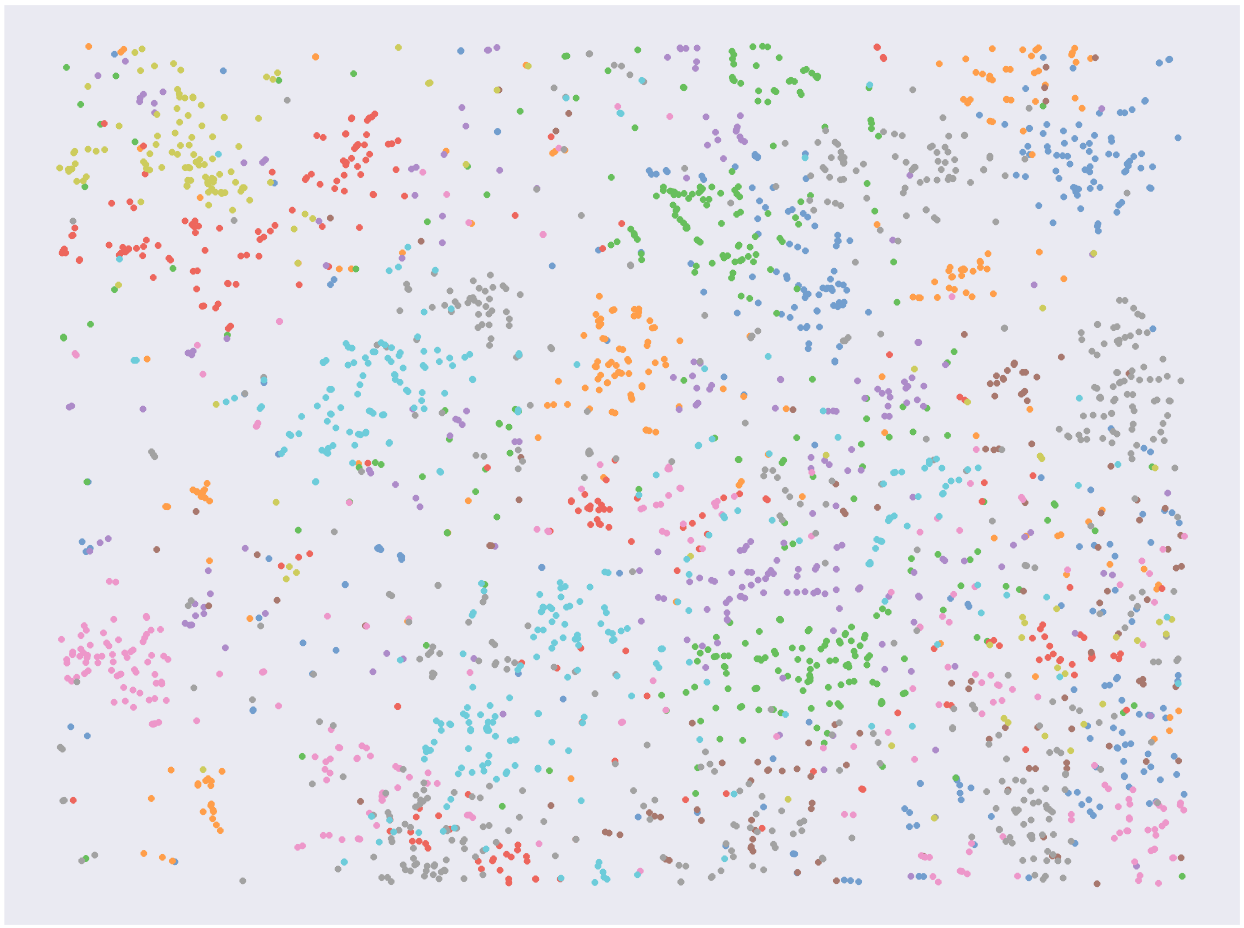}
				\end{minipage}
			}
			\subfigure[Label Smoothing (LS)]{
				\begin{minipage}{0.31\linewidth}
					\includegraphics[width=\linewidth]{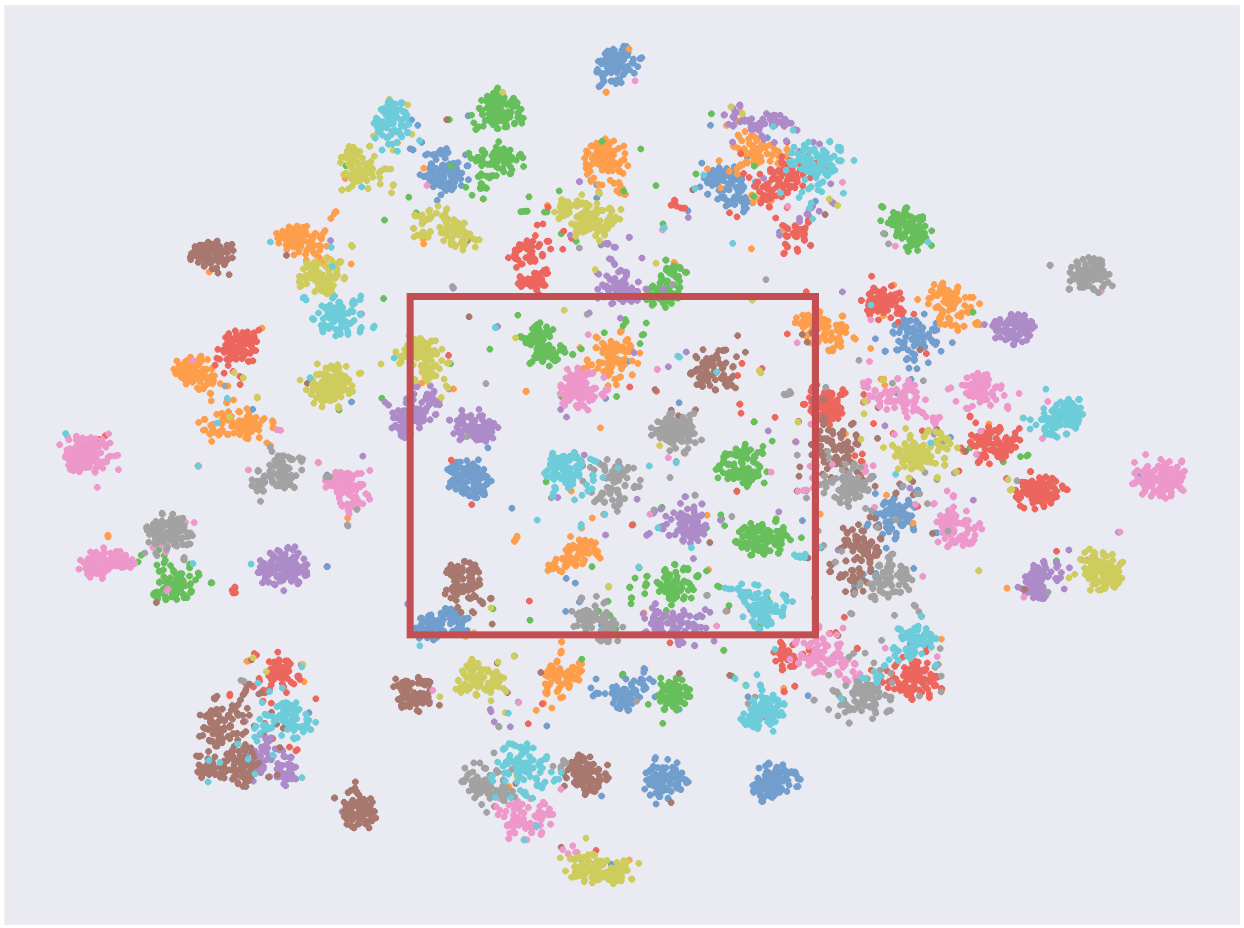}\\
					\includegraphics[width=\linewidth]{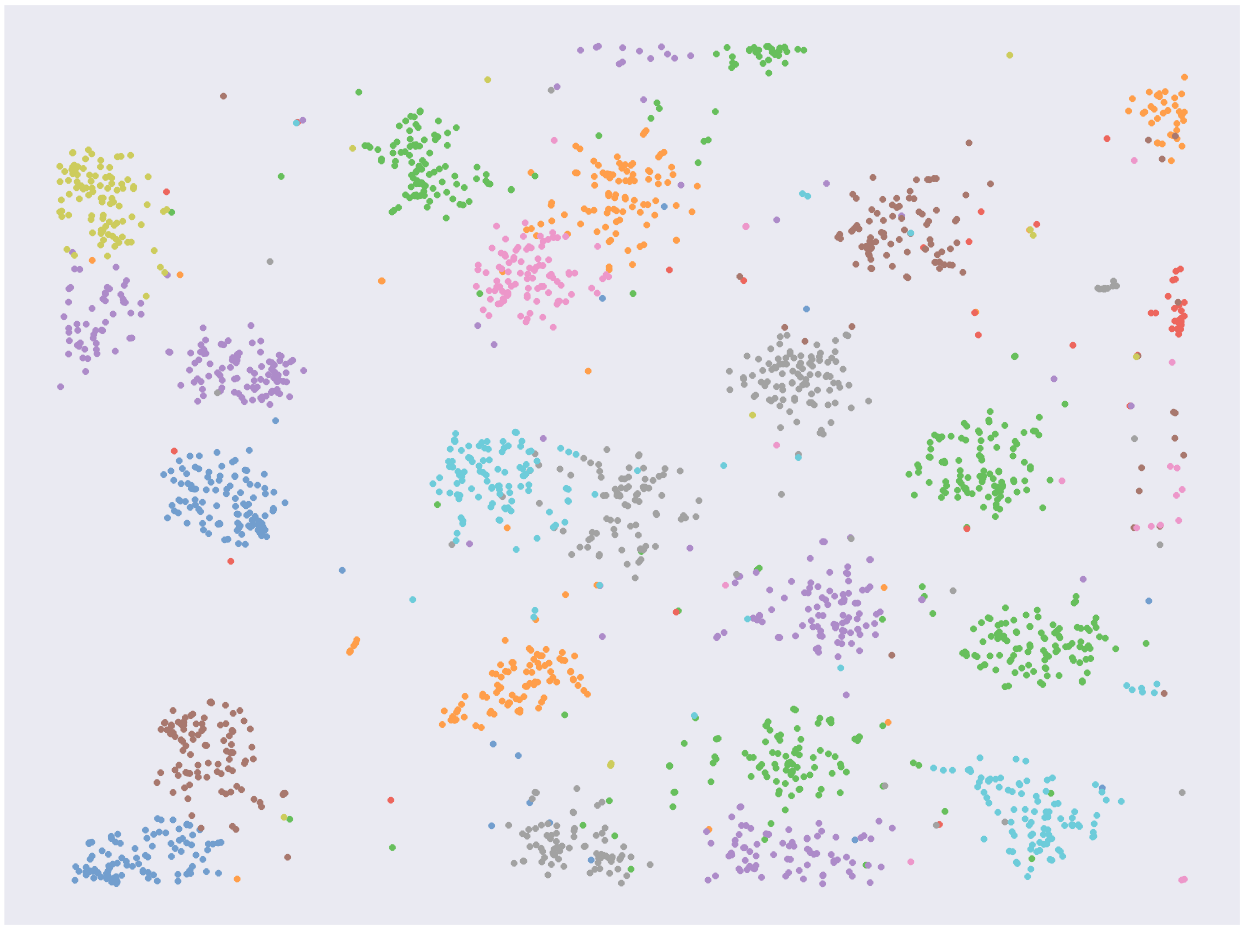}
				\end{minipage}
			}
			\subfigure[OLS (Ours)]{
				\begin{minipage}{0.31\linewidth}
					\includegraphics[width=\linewidth]{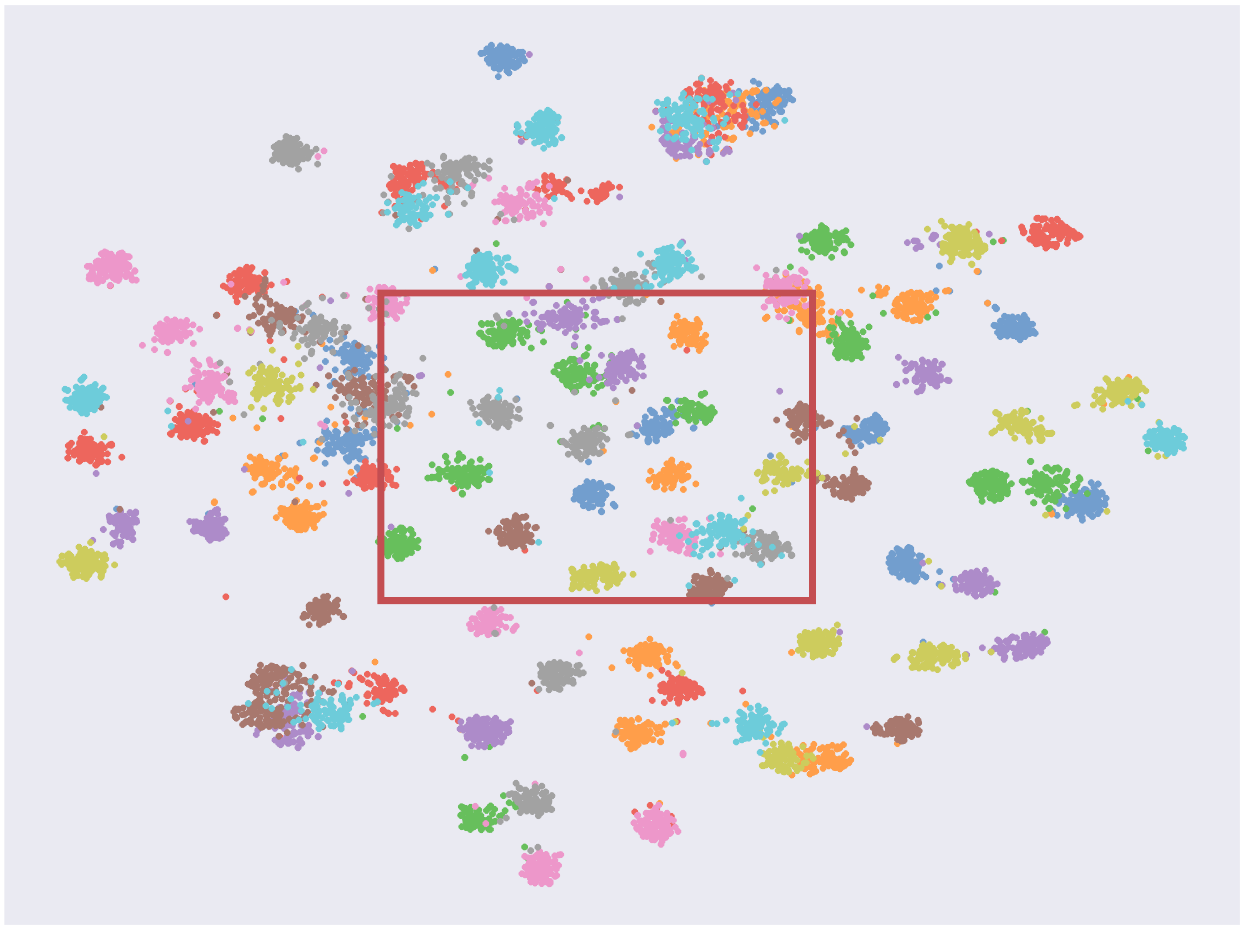}\\
					\includegraphics[width=\linewidth]{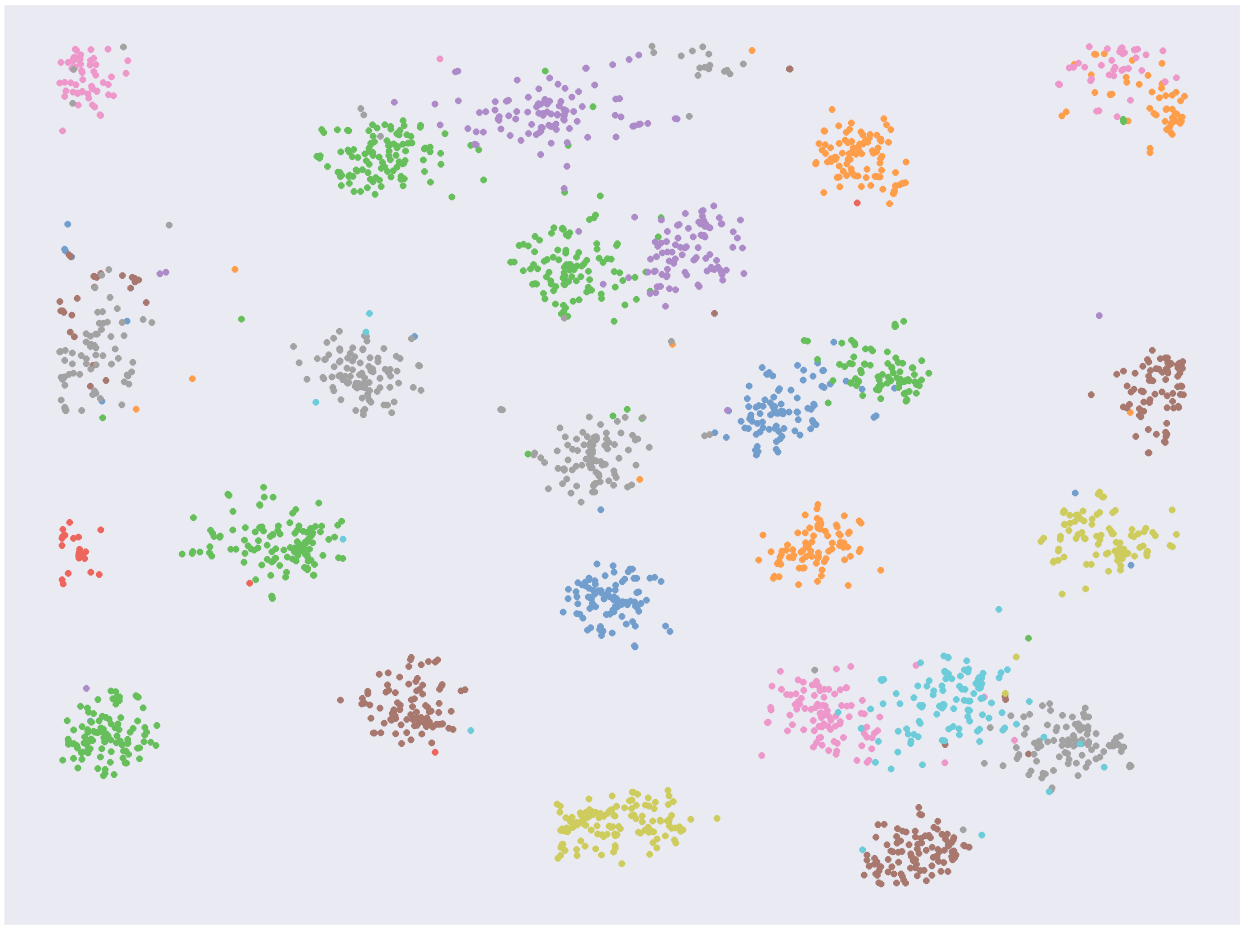}
				\end{minipage}
			}		
			\vskip -0.1in
			\caption{Visualization of the penultimate layer representations of ResNet-56 on CIFAR-100 training set
				using t-SNE~\cite{tsne}.
				Note that we use the same color for every 10 classes.
				We visualize the representations of all 100 classes (\textbf{top}).
				We zoom the patch in red boxes for better visualization. (\textbf{bottom}).  
			} \label{fig:embedding}
		\end{minipage}
	\end{small}
	\vskip -0.1in
\end{figure*}

In the $t_{th}$ training epoch, we also use the predicted probabilities 
of the input samples to update $S^{t}_{y_i}$, which will be utilized 
to supervise the model training in the $t+1$ epoch. 
At the beginning of the $t_{th}$ training epoch, 
we initialize the soft label $S^t$ as a zero matrix.
When an input sample $(\bm{x_i}, y_i)$ is correctly 
classified by the model, we utilize its predicted score $p(\bm{x_i})$ 
to update the $y_i$ column in $S_{t}$, which can be formulated as
\begin{equation} 
	S_{y_i, k}^t = S_{y_i, k}^t + p(k|\bm{x_i}), \label{eq:accumulate}
\end{equation}
where $k \in \{1, \cdots, K\}$, indexing the soft label $S_{y_i}^t$.
At the end of the $t_{th}$ training epoch, we normalize 
the cumulative $S^t$ column by column as represented by 
\begin{equation}
	S_{y_i, k}^{t} \leftarrow \frac{S_{y_i, k}^{t}}{\sum_{l=1}^{K}S_{y_i,l}^{t}}.
\end{equation}
We can now obtain the normalized soft label $S^t$ for all $K$ 
categories, which will be used to supervise the model 
at the next training epoch.
Notice that we cannot obtain the soft label at the first epoch.
Thus, we use the uniform distribution to initialize each column 
in $S^0$.
More details for the proposed approach are described 
in Algorithm~\ref{alg:ols}.

\paragraph{Discussion} 
The soft label $S^{t-1}_{y_i, k}$ generated from the $t-1$ epoch can 
be denoted as  
\begin{equation}
S^{t-1}_{y_i, k} = \frac{1}{N}\sum_{j=1}^N p^{t-1}(k | \bm{x_j}), 
\end{equation}
where $N$ denotes the number of correctly predicted samples 
with label $y_i$. 
$p^{t-1}(k | \bm{x_j})$ is the output probability of category $k$ 
when input $x_j$ to the network at the $t-1$ epoch. 
Then \eqref{eq:softce} can be rewritten as
\begin{equation}
\begin{aligned}
    L_{soft} = -\sum_{k=1}^K \frac{1}{N} \sum_{j=1}^N p^{t-1}(k|\bm{x_j}) \cdot \log{p(k|\bm{x_i})} \\ 
    = -\frac{1}{N} \sum_{j=1}^N \sum_{k=1}^K p^{t-1}(k|\bm{x_j}) \cdot \log p(k|\bm{x_i}).
 \end{aligned} \label{eq:intraclass}
 \end{equation}
This equation indicates that all correctly classified samples $\bm{x_j}$ will impose 
a constraint to the current sample $\bm{x_i}$.
The constrain encourages the samples belonging to the same category to be much closer.
To give a more intuitive explanation, we utilize t-SNE~\cite{tsne} 
to visualize the penultimate layer representations of ResNet-56 
on CIFAR-100 trained with the hard label, LS, and OLS,
respectively.
\figref{fig:embedding} shows that 
our proposed method provides a more recognizable difference
between representations of different classes
and tighter intra-class representations.

Besides, our method does not have the problem of training divergence in the early stages of training.
This is because we use a uniform distribution as the soft label
in the first epoch of training,
which is equivalent to the vanilla label smoothing.
In the entire training process afterwards,
we only accumulate correct predictions,
which guarantees the correctness of
the generated soft labels.

\begin{table*}[t]
  \centering
  \renewcommand{\tabcolsep}{1.5mm} 
  \caption{
    Comparison between our method and the state-of-the-art approaches.
    We run each method three times on CIFAR-100 and compute the mean and standard deviation of the Top-1 Error (\%).
    Best results are highlighted in \textbf{bold}.
  }\label{tab:cifar}
  \begin{tabular}{lccccc}
    \toprule
    Method & ResNet-34 & ResNet-50 & ResNet-101 & ResNeXt29-2x64d  & ResNeXt29-32x4d \\
    \midrule
    {Hard Label} 
    & 20.62 $\pm$ 0.21 
    & 21.21 $\pm$ 0.25
    & 20.34 $\pm$ 0.40 
    & 20.92 $\pm$ 0.52 
    & 20.85 $\pm$ 0.17 \\
    {Bootsoft~\cite{bootstrap}}
    & 21.65 $\pm$ 0.13 
    & 21.25 $\pm$ 0.67 
    & 20.37 $\pm$ 0.07 
    & 21.20 $\pm$ 0.13 
    & 20.86 $\pm$ 0.24 \\
    {Boothard~\cite{bootstrap}} 
    & 22.58 $\pm$ 0.02 
    & 20.81 $\pm$ 0.13
    & 21.46 $\pm$ 0.22
    & 21.00 $\pm$ 0.10 
    & 21.47 $\pm$ 0.59 \\
    {Disturb Label~\cite{Xie2016disturb}} 
    & 20.91 $\pm$ 0.30 
    & 22.12 $\pm$ 0.51
    & 20.99 $\pm$ 0.12
    & 21.64 $\pm$ 0.24
    & 21.69 $\pm$ 0.06\\
    {SCE~\cite{sl}}
    & 22.86 $\pm$ 0.08
    & 22.12 $\pm$ 0.11 
    & 22.60 $\pm$ 0.64
    & 23.07 $\pm$ 0.28
    & 22.96 $\pm$ 0.09 \\
    {LS~\cite{christian2016rethinking}} 
    & 20.94 $\pm$ 0.08
    & 21.20 $\pm$ 0.25 
    & 20.12 $\pm$ 0.02
    & 20.34 $\pm$ 0.24
    & 19.56 $\pm$ 0.18 \\
    {Pairwise Confusion~\cite{dubey2018pairwise}} 
    & 22.91 $\pm$ 0.04 
    & 23.09 $\pm$ 0.53 
    & 22.73 $\pm$ 0.39 
    & 21.55 $\pm$ 0.11 
    & 21.74 $\pm$ 0.04 \\
    {Xu \etal ~\cite{xu2019data}}
    & 22.65 $\pm$ 0.09 
    & 22.05 $\pm$ 0.43 
    & 21.70 $\pm$ 0.77 
    & 22.81 $\pm$ 0.08 
    & 23.14 $\pm$ 0.04 \\
    {OLS}
    & \bf20.04 $\pm$ 0.11
    & \bf20.65  $\pm$ 0.14
    & \bf19.66  $\pm$ 0.15 
    & \bf18.81  $\pm$ 0.45
    & \bf18.79  $\pm$ 0.20 \\
    \midrule
    {BYOT~\cite{zhang2019byot}} 
    & 20.41 $\pm$ 0.10 
    & 19.20 $\pm$ 0.30
    & 18.51 $\pm$ 0.49 
    & 19.69 $\pm$ 0.12
    & 20.33 $\pm$ 0.19 \\
    {BYOT~\cite{zhang2019byot} + OLS} 
    & \bf19.44 $\pm$ 0.09
    & \bf18.15 $\pm$ 0.21 
    & \bf18.14 $\pm$ 0.08
    & \bf18.29 $\pm$ 0.20 
    & \bf19.25 $\pm$ 0.29 \\
    \bottomrule
  \end{tabular}
\end{table*}

\section{Discussion}

\textbf{Comparison with Tf-KD~\cite{yuan2020revisiting}.}
The output of the teacher model in Label Smoothing~\cite{christian2016rethinking} is a uniform distribution.
Yuan~\etal~\cite{yuan2020revisiting} argue that this uniform distribution could not
reflect the correct class information,
so they propose a teacher-free knowledge distillation method,
called Tf-KD$_{reg}$.
They design a teacher with correct class information.
The output of the teacher model can be denoted as:
\begin{equation}
u(k) = \begin{cases}
a & \text{ if } k=c \\
\frac{1-a}{K-1}  & \text{ if } k \ne  c
\end{cases} \label{eq:tfkd}
\end{equation}
where $u(k)$ is the hand-designed distribution,
$c$ is the correct class and $K$ is the number of classes.
They set the hyper-parameter $a > 0.9$.
Although both this distribution and our method could contain the correct class information,
the hand-designed distribution of Tf-KD$_{reg}$ is still uniform distribution among non-target classes.
The distribution of Tf-KD~\cite{yuan2020revisiting} still does not imply similarities between classes.
On the contrary,
our motivation is to find a non-uniform distribution that can reflect the relationship between classes.
Hinton~\etal~\cite{hinton15KD}, Borns Again Network~\cite{tommaso2018ban} and Tf-KD~\cite{yuan2020revisiting} have emphasized this view
that knowledge distillation benefits from the similarities among classes implied in the output of the teacher model.
We conduct experiments on four fine-grained datasets as shown in~\tabref{tab:fine}.
Our method benefits from the similarities between classes,
so it can perform better than Tf-KD~\cite{yuan2020revisiting}.

\begin{table*}[!htp]
	\centering
	\caption{The Top-1 Error of model ensemble. We integrate 6, 10, 15, and 20 models trained at different epochs, respectively.
		The models are selected uniformly from all training epochs (300 epochs).} \label{tab:ensem}
	\renewcommand{\tabcolsep}{4mm}
	\begin{tabular}{c|c|cccc}
		\toprule
		Method     & 1 Model & 6 Models & 10 Models & 15 Models & 20 Models \\
		\midrule
		HardLabel  & 26.41  & 26.07  & 25.93 & 25.87 & 25.88 \\
		LS~\cite{christian2016rethinking}  & 26.37 & 25.30 & 25.11 & 24.97  & 24.96 \\
		OLS (ours) & 25.27  & 24.52   & 24.22 & 24.10 & 23.91  \\
		\bottomrule
	\end{tabular}
\end{table*}

\textbf{Connection with the model ensemble.}
Integrating models trained at different epochs
is an effective and cost-saving ensemble method.
The way to integrate the outputs of models trained at
different epochs is described as follows:\\
\begin{equation}
\begin{aligned}
z_{i} =\frac{1}{||T||} \sum_{t\in T} softmax(W(x_{i}|\theta_{t})),
\end{aligned}
\end{equation}
where $z_{i}$ denotes the ensemble predictions,
$T$ denotes the set of selected models in different epochs,
$W$ denotes the network,
$\theta_{t}$ denotes the network parameters in t-th
epoch and $x_{i}$ denotes the input sample.
Both our method and the model ensemble utilize the knowledge from 
different training epochs.
The model ensemble averages the outputs of models 
at different epochs to make predictions.
However,
different from the ensemble method,
our method utilize the knowledge from the previous epoch to help
the learning in the current epoch.
Specifically,
our method generates the soft labels in one training epoch,
and the soft labels are used to supervise the network training.
It is worth noting that our method does not conflict
with this ensemble strategy.
To verify this point,
we conduct experiments on CIFAR-100 using ResNet-56.
We apply the same experimental setup described in the~\secref{sec:experiments}.
Experimental results are shown in~\tabref{tab:ensem}.
For all methods,
we apply the same ensemble strategy.
We select models uniformly from the whole training schedule (300 epochs).
We choose 6, 10, 15 and 20 models for ensemble respectively.
In \tabref{tab:ensem}, our method achieves 25.27\% Top-1 Error. 
When our method equipping with the ensemble method, the performance is further 
improved by a large margin (`20 Models': 23.91\%).
The experiments show that there is no conflict between our method 
and the model ensemble.

\section{Experiments} \label{sec:experiments}
In \secref{sec:generalimagerecognition}, we first present and analyze 
the performance of our approach on CIFAR-100, ImageNet, and some 
fine-grained datasets.
Then,
we test the tolerance to symmetric noisy labels in \secref{sec:tolerance}
and robustness to adversarial attacks in \secref{sec:robustness},
respectively.
In \secref{sec:objectdet},
we apply our OLS to object detection.
Moreover, in \secref{sec:ablation}, we conduct extensive ablation experiments 
to analyze the settings of our method.	
All the experiments are implemented based on PyTorch \cite{paszke2017automatic}
and Jittor \cite{hu2020jittor}.

\subsection{General Image Recognition} \label{sec:generalimagerecognition}
\textbf{CIFAR Classification.} \label{para:cifar}
First, we conduct experiments on CIFAR-100 dataset to compare 
our OLS with other related methods, including regularization 
methods on labels (Bootstrap~\cite{bootstrap},
Disturb Label~\cite{Xie2016disturb},
Symmetric Cross Entropy~\cite{sl},
Label Smoothing~\cite{christian2016rethinking}
and Pairwise Confusion~\cite{dubey2018pairwise})
and self-knowledge distillation methods
(Xu \etal~\cite{xu2019data} and BYOT~\cite{zhang2019byot}).
For a fair comparison with them, we keep the same experimental setup 
for all methods.
Specifically, we train all the models for 300 epochs with a batch size of 128. 
The learning rate is initially set to 0.1 and decays at the 
150$th$ and 225$th$ epoch by a factor of 0.1, respectively.
For other hyper-parameters in different methods, we keep 
their original settings.
Additionally, for a fair comparison with BYOT~\cite{zhang2019byot} 
and Xu \etal~\cite{xu2019data}, we remove the feature-level supervision in 
them and only use the class labels to supervise models.

\tabref{tab:cifar} shows the classification 
results of each method based on different network architectures.
It can be seen that our method significantly improves 
the classification performance on both lightweight 
and complex models, which indicates its robustness 
to different networks.
Since BYOT~\cite{zhang2019byot} is learned with deep supervision,
it performs better on deeper models, like ResNet-50 and
ResNet-101, than our method.
However, our method can be easily plugged into BYOT~\cite{zhang2019byot}
and achieves better results than BYOT on deeper models.
In addition, comparing to LS~\cite{christian2016rethinking},
our method achieves stable improvement on different models.
Especially, our method outperforms LS by about 1.5\%
on ResNeXt29-2x64d.
We argue that the performance gain owes to the useful relationships 
among categories discovered by our soft labels.
In \secref{sec:ablation}, we will further analyze the importance 
of building relationships among categories. 

\textbf{ImageNet Classification.} 
We also evaluate our method on a large-scale dataset, ImageNet.
It contains 1K categories with a total of 1.2M training images and 
50K validation images.
Specifically, we use the SGD optimizer 
to train all the models for 250 epochs with a batch size of 256.
The learning rate is initially set to 0.1 and decays 
at the 75$th$, 150$th$, and 225$th$ epochs, respectively.
We report the best performance of each method. 

The classification performance on ImageNet dataset
is shown in \tabref{tab:imagenet}.
Applying our OLS to ResNet-50 achieves 22.28\% Top-1 Error, 
which is better than the result with LS~\cite{christian2016rethinking} 
by 0.54\%.  
Additionally, ResNet-101 with our OLS can achieve 20.85\% top-1 
error, which improves ResNet-101 by 1.02\% and ResNet-101 with
LS by 0.42\%, respectively. 
This demonstrates that our OLS still performs well on 
the large-scale dataset.
Moreover, we explore the combination of our method with other strategies, i.e., 
data augmentation (CutOut~\cite{devries2017cutout}) and self-distillation (BYOT~\cite{zhang2019byot}).
In \tabref{tab:imagenet}, 
we observe the combination with them brings extra performance gains 
to ResNet50 and ResNet101.
Our OLS can be utilized as a plug-in regularization module, which is 
easy to be combined with other methods.

\begin{table}
	\centering
  \renewcommand{\tabcolsep}{4mm} 
	\caption{Classification results on ImageNet. $\ddagger$ denotes the results reported in Tf-KD~\cite{yuan2020revisiting}.
	}\label{tab:imagenet}
	\begin{tabular}{lcc}
		\toprule
		{Model}  &\makecell{Top-1\\{Error(\%)}} & \makecell{{Top-5}\\Error(\%)} \\
		\midrule 
		ResNet-50 & 24.23$^\ddagger$ & - \\
		ResNet-50 + LS~\cite{christian2016rethinking} & 23.62$^\ddagger$ & - \\
		ResNet-50 + Tf-KD$_{self}$~\cite{yuan2020revisiting} & 23.59$^\ddagger$ & - \\
		ResNet-50 + Tf-KD$_{reg}$~\cite{yuan2020revisiting} & 23.58$^\ddagger$ & - \\
		\midrule
		ResNet-50 &23.68 & 7.05 \\
		ResNet-50 + Bootsoft~\cite{bootstrap} & 23.49 & 6.85 \\
		ResNet-50 + Boothard~\cite{bootstrap} & 23.85 & 7.07 \\
		ResNet-50 + LS~\cite{christian2016rethinking} & 22.82 & 6.66 \\
		ResNet-50 + CutOut~\cite{devries2017cutout} & 22.93 & 6.66 \\
		ResNet-50 + Disturb Label~\cite{Xie2016disturb} & 23.59 & 6.90 \\
		ResNet-50 + BYOT~\cite{zhang2019byot} & 23.04 & 6.51 \\
		ResNet-50 + OLS & 22.28 & 6.39 \\
		ResNet-50 + CutOut~\cite{devries2017cutout} + OLS & 21.98 & \textbf{6.18} \\
		ResNet-50 + BYOT~\cite{zhang2019byot} + OLS & \textbf{21.88} & 6.27 \\
		\midrule
		ResNet-101 & 21.87 & 6.29 \\
		ResNet-101 + LS~\cite{christian2016rethinking} &  21.27 & 5.85 \\
		ResNet-101 + CutOut~\cite{devries2017cutout} & 20.72 & 5.51 \\
		ResNet-101 + OLS & 20.85 & 5.50 \\
		ResNet-101 + CutOut~\cite{deng2009imagenet} + LS~\cite{christian2016rethinking} & 20.47 & 5.51\\
		ResNet-101 + CutOut~\cite{deng2009imagenet} + OLS & \textbf{20.25} & \textbf{5.42} \\
		\bottomrule
	\end{tabular}
\end{table}
\begin{table}[tp!]
  \centering
  \setlength{\tabcolsep}{1mm}
  \caption{Detailed information of the fine-grained datasets.}
  \begin{tabular}{lcccc}
      \toprule
      Dataset & Categories & Training Samples & Test Samples  \\
      \midrule 
      CUB-200-2011~\cite{WahCUB_200_2011} & 200 & 5994 & 5794 \\
      Flowers-102~\cite{nilsback2008automated} & 102 & 2040 & 6149 \\
      Cars~\cite{krause20133d} & 196 & 8144 & 8041 \\
      Aircrafts~\cite{maji2013fine} & 90 & 6667 & 3333 \\
      \bottomrule
  \end{tabular}  \label{tab:info}
\end{table}

\begin{table*}[th!]
	\renewcommand{\tabcolsep}{1mm}
	\centering
	\caption{The Top-1 and Top-5 Error(\%) of different architectures on 
		fine-grained classification datasets. 
		All results are averaged over three runs.
		The $\triangle$ denotes the average improvement relative to Hard Label on all datasets and backbones.
	}\label{tab:fine}
	\resizebox{\textwidth}{!}{
	\begin{tabular}{lccccccccc}
		\toprule
		\multirow{2}*{Dataset} &
		\multirow{2}*{\makecell{Backbones}} &
		\multicolumn{2}{c}{Hard Label} &
		\multicolumn{2}{c}{LS~\cite{christian2016rethinking}} &
		\multicolumn{2}{c}{Tf-KD~\cite{yuan2020revisiting}} &
		\multicolumn{2}{c}{OLS} \\
		\cmidrule(lr{0.25em}){3-4} 
		\cmidrule(lr{0.25em}){5-6}
		\cmidrule(lr{0.25em}){7-8}
		\cmidrule(lr{0.25em}){9-10}
		~  & ~ &
		\makecell{Top-1 Err(\%)} & 
		\makecell{Top-5 Err(\%)} & 
		\makecell{Top-1 Err(\%)} & 
		\makecell{Top-5 Err(\%)} &
		\makecell{Top-1 Err(\%)} &
		\makecell{Top-5 Err(\%)} &
		\makecell{Top-1 Err(\%)} &
		\makecell{Top-5 Err(\%)}\\
		\midrule
		{CUB-200-2011~\cite{WahCUB_200_2011}}  & \multirow{4}*{\makecell{ResNet-50~\cite{he2016deep}}} &
		$19.19\pm 0.22$ & $5.00\pm 0.25$ &
		$18.11\pm 0.14$ & $4.88\pm 0.08$ &
		$19.04\pm 0.23$ & $4.92\pm 0.16$ &
		$17.53\pm 0.09$ & $4.01\pm 0.27$ \\
		{Flowers-102~\cite{nilsback2008automated}} & ~ &
		$9.31\pm 0.19$ & $2.43\pm 0.14$ &
		$7.58\pm 0.07$ & $1.93\pm 0.03$ & 
		$8.70\pm 0.45$ & $2.46\pm 0.09$ &
		$7.14\pm 0.14$ & $1.55\pm 0.07$ \\
		Cars~\cite{krause20133d}  & ~ &
		$9.58\pm 0.19$ & $1.79\pm 0.01$ & 
		$8.32\pm 0.09$ & $1.57\pm 0.03$ &
		$8.65\pm 0.16$ & $1.46\pm 0.10$ &
		$7.46\pm 0.01$ & $0.92\pm 0.04$ \\
		Aircrafts~\cite{maji2013fine}  & ~ &
		$11.88\pm 0.11$ & $3.86\pm 0.13$ &
		$9.92\pm 0.07$ & $3.73\pm 0.12$ &
		$10.55\pm 0.22$ & $3.34\pm 0.21$ &
		$9.19\pm 0.12$ & $2.60\pm 0.03$ \\

				\midrule
		{CUB-200-2011~\cite{WahCUB_200_2011}}  & \multirow{4}*{\makecell{MobileNetv2~\cite{sandler18mobile}}} &
		$22.24\pm 0.33$ & $6.61\pm 0.21$ &
		$21.33\pm 0.29$ & $7.05\pm 0.09$ &
		$22.36\pm 0.27$ & $6.41\pm 1.47$ &
		$20.05\pm 0.11$ & $5.08\pm 0.12$\\
		{Flowers-102~\cite{nilsback2008automated}} & ~ &
		$8.97\pm 0.09$ & $2.51\pm 0.19$ &
	    $8.06\pm 0.35$ & $2.46\pm 0.08$ & 
	    $8.05\pm 0.14$ & $2.23\pm 0.13$ &
		$7.27\pm 0.17$ & $1.77\pm 0.10$ \\
		Cars~\cite{krause20133d}  & ~ &
		$11.71\pm 0.13$ & $2.29\pm 0.12$ &
	    $10.17\pm 0.07$ & $2.33\pm 0.05$ & 
	    $10.57\pm 0.09$ & $2.14\pm 0.04$ &
		$9.25\pm 0.05$ & $1.33\pm 0.02$ \\
		Aircrafts~\cite{maji2013fine}  & ~ &
		$13.16\pm 0.33$ & $4.15\pm 0.19$ &
	    $12.05\pm 0.29$ & $4.08\pm 0.17$ & 
	    $11.95\pm 0.27$ & $4.04\pm 0.10$ &
		$10.53\pm 0.25$ & $2.96\pm 0.15$ \\

						\midrule
		{CUB-200-2011~\cite{WahCUB_200_2011}}  & \multirow{4}*{\makecell{EfficientNet-b7~\cite{efficientNet}}} &
		$18.44\pm 0.15$ & $5.07\pm 0.13$ &
		$17.40\pm 0.14$ & $5.02\pm 0.03$ &
		$20.24\pm 0.09$ & $6.33\pm 0.21$ &
		$16.21\pm 0.24$ & $3.34\pm 0.02$\\
		{Flowers-102~\cite{nilsback2008automated}} & ~ &
		$9.50\pm 0.07$ & $2.04\pm 0.07$ &
	    $9.42\pm 0.34$ & $2.34\pm 0.13$ & 
	    $8.58\pm 0.37$ & $2.07\pm 0.10$ &
		$8.16\pm 0.12$ & $1.63\pm 0.15$ \\
		Cars~\cite{krause20133d}  & ~ &
		$9.24\pm 0.22$ & $1.84\pm 0.13$ &
	    $8.42\pm 0.08$ & $1.76\pm 0.07$ &
	    $9.52\pm 0.01$ & $1.64\pm 0.01$ & 
		$7.53\pm 0.13$ & $0.97\pm 0.02$ \\
		Aircrafts~\cite{maji2013fine}  & ~ &
		$11.61\pm 0.37$ & $3.72\pm 0.20$ &
	    $9.60\pm 0.15$ & $3.62\pm 0.13$ & 
	    $9.45\pm 0.49$ & $2.01\pm 0.04$ &
		$8.83\pm 0.19$ & $2.71\pm 0.12$ \\

				\midrule
		{CUB-200-2011~\cite{WahCUB_200_2011}}  & \multirow{4}*{\makecell{SAN-15~\cite{zhao2020exploring}}} &
		$19.05\pm 0.39$ & $5.37\pm 0.25$ &
		$17.54\pm 0.30$ & $5.43\pm 0.19$ &
		$19.88\pm 0.17$ & $5.81\pm 0.03$ &
		$17.28\pm 0.14$ & $4.08\pm 0.07$\\
		{Flowers-102~\cite{nilsback2008automated}} & ~ &
		$7.85\pm 0.29$ & $1.78\pm 0.21$ &
	    $8.08\pm 0.34$ & $1.95\pm 0.15$ & 
	    $7.87\pm 0.43$ & $1.91\pm 0.28$ &
		$7.09\pm 0.18$ & $1.56\pm 0.12$ \\
		Cars~\cite{krause20133d}  & ~ &
		$9.23\pm 0.07$ & $1.78\pm 0.02$ &
	    $8.55\pm 0.15$ & $1.87\pm 0.04$ & 
	    $8.98\pm 0.07$ & $1.76\pm 0.14$ &
		$7.55\pm 0.14$ & $1.08\pm 0.07$ \\
		Aircrafts~\cite{maji2013fine}  & ~ &
		$11.31\pm 0.13$ & $3.79\pm 0.08$ &
	    $9.96\pm 0.09$ & $3.45\pm 0.14$ & 
	    $10.77\pm 0.03$ & $4.18\pm 0.08$ &
		$9.43\pm 0.08$ & $2.95\pm 0.09$ \\

        \midrule
        {Average Improvements ($\triangle$)} & ~ & 0.00 & 0.00 &
        1.11 \textcolor{red}{$\uparrow$} & 0.02 \textcolor{red}{$\uparrow$} &
        0.44 \textcolor{red}{$\uparrow$} & 0.19 \textcolor{red}{$\uparrow$}
        & \textbf{2.00} \textcolor{red}{$\uparrow$} & \textbf{0.96} \textcolor{red}{$\uparrow$} \\
		\bottomrule
	\end{tabular}
	}
\end{table*}

\begin{table*}[t]
	\centering
	\renewcommand{\tabcolsep}{3.2 mm} 
	\caption{The classification performance of different methods under
		different noisy rates.
		We run each method three times under different noisy rates and compute the mean and standard deviation of the Top-1 Error(\%).
		The best two results are in \textbf{bold}.
	}\label{tab:noisylabels}
	\begin{tabular}{lccccc}
		\toprule
		Method/Noise Rate & 0\% & 20\% & 40\% & 60\% & 80\% \\
		\midrule
		{Hard Label} 
		& 26.81 $\pm$ 0.36 & 37.75 $\pm$ 0.50 & 47.07 $\pm$ 1.08
		& 62.06 $\pm$ 0.62 & 81.56 $\pm$ 0.42 \\
		{Bootsoft~\cite{bootstrap}} 
		& 27.28 $\pm$ 0.35 & 37.99 $\pm$ 0.43 & 46.96 $\pm$ 0.33 
		& 63.76 $\pm$ 0.85 & 80.32 $\pm$ 0.33 \\
		{Boothard~\cite{bootstrap}} 
		& \bf26.02 $\pm$ 0.22 & 36.21 $\pm$ 0.29 & 42.73 $\pm$ 0.16 
		& 54.95 $\pm$ 2.20 & 81.20 $\pm$ 1.26 \\
		{Symmetric Cross Entropy~\cite{sl}} 
		& 28.97 $\pm$ 0.31 & 38.40 $\pm$ 0.12 & 46.97 $\pm$ 0.65 
		& 62.13 $\pm$ 0.55 & 82.66 $\pm$ 0.10 \\
		{Ren \etal \cite{Ren2018reweight}} 
		& 38.38 $\pm$ 0.35 & 43.74 $\pm$ 1.21 & 49.83 $\pm$ 0.53
		& 57.65 $\pm$ 0.98 & \bf73.04 $\pm$ 0.15 \\
		{MetaWeightNet~\cite{metaweight}} 
		& 29.51 $\pm$ 0.51 & 35.06 $\pm$ 0.48 & 43.58 $\pm$ 0.93 & 56.15 $\pm$ 0.60 & 87.25 $\pm$ 0.22 \\
		{Arazo \etal \cite{unsupervised}} 
		& 33.80 $\pm$ 0.10 & \bf33.91 $\pm$ 0.38 & \bf40.87 $\pm$ 1.49 & \bf52.91 $\pm$ 1.81 & 83.92 $\pm$ 0.19 \\
		{PENCIL~\cite{yi2019probabilistic}} & 29.36 $\pm$ 0.35 & 36.33 $\pm$ 0.15 & 43.55 $\pm$ 0.08 & 57.49 $\pm$ 1.05 & 79.24 $\pm$ 0.11 \\
		{Han~\etal~\cite{han2019deep}} & 32.07 $\pm$ 0.36 & 35.08 $\pm$ 0.19 & 44.39 $\pm$ 0.23 & 62.50 $\pm$ 0.61  & 80.39 $\pm$ 0.16 \\
		\midrule
		{LS \cite{christian2016rethinking}} 
		& 26.37 $\pm$ 0.41 & 35.48 $\pm$ 0.61 & 43.99 $\pm$ 1.04 
		& 59.51 $\pm$ 0.80 & 80.36 $\pm$ 0.90 \\
		{OLS}  
		& \bf25.24 $\pm$ 0.18 & \bf32.67 $\pm$ 0.14 & \bf38.86 $\pm$ 0.13 
		& \bf50.04 $\pm$ 0.14 & \bf78.22 $\pm$ 1.01 \\
		\bottomrule
	\end{tabular}
\end{table*}

\textbf{Fine-grained Classification.}
The fine-grained image classification task~\cite{iscen2015comp,zhang2017fine,zhang2016weak,shi2019fine,shu2016image}
focuses on distinguishing subordinate categories
within entry-level categories~\cite{dubey2018pairwise,peng2018object,zheng2020learn,lin2018bilinear}.
We conduct experiments on four fine-grained image
recognition datasets, 
including CUB-200-2011~\cite{WahCUB_200_2011},
Flowers-102~\cite{nilsback2008automated},
Cars~\cite{krause20133d}
and Aircrafts~\cite{maji2013fine}
, respectively.
In \tabref{tab:info}, we present the details of these datasets.
For all experiments, we keep the same experimental setup.
Specifically, we use SGD as the optimizer and train all models for 
100 epochs.
The initial learning rate is set as 0.01 and it decays at the 45$th$ 
epoch and 80$th$ epoch, respectively.
In \tabref{tab:fine}, we report the average
Top-1 Error(\%) and Top-5 Error(\%) of three runs.
Experiment results demonstrate that
OLS can also improve classification performance 
on the fine-grained datasets, which indicates 
our soft labels can benefit fine-grained category 
classification.

\subsection{Tolerance to Noisy Labels}\label{sec:tolerance}

As demonstrated in~\cite{sl,xiao2015learning},
there exist noisy (incorrect) labels in datasets,
especially those obtained from webs.
Due to the powerful fitting ability of DNNs,
they can still fit noisy labels easily~\cite{ZhangBHRV17}.
But this is harmful for the generalization of DNNs.
To reduce such damage to the generalization ability of DNNs,
researchers have proposed many methods,
including weighting the samples~\cite{Ren2018reweight,metaweight}
and inferring the real labels of the noisy samples~\cite{bootstrap,unsupervised}.
We notice that our method can improve the performance of DNNs
on noisy labels by reducing the fitting to noisy samples. 
We conduct experiments on CIFAR-100 to verify the regularization 
capability of our method on noisy data.

We follow the same experimental settings as in~\cite{sl,unsupervised}.
We randomly select a certain number of samples according to 
the noisy rate and flip the labels of these samples to the 
wrong labels uniformly (symmetric noise) before training.
Since both Ren \etal~\cite{Ren2018reweight} and 
MetaWeightNet~\cite{metaweight} need to
split a part of the clean validation set from the training set,
we keep their default optimal number of samples 
in the validation set.

\begin{figure*}[!ht]
	\begin{small}
		\centering
		\subfigure{ 
			\begin{overpic}[width=0.31\linewidth, tics=5]{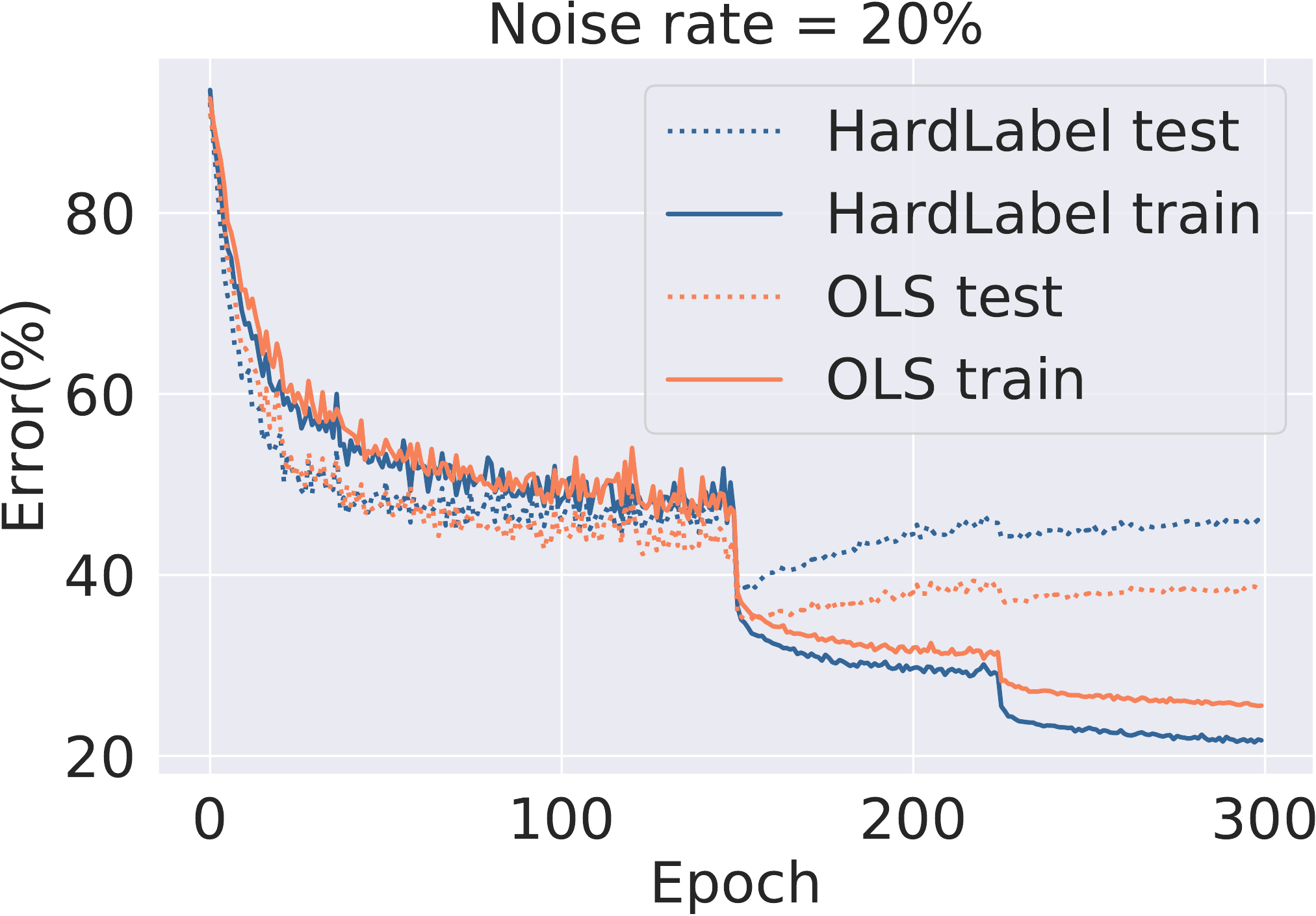}
			\end{overpic}
		}
		\subfigure{ 
			\begin{overpic}[width=0.31\linewidth, tics=5]{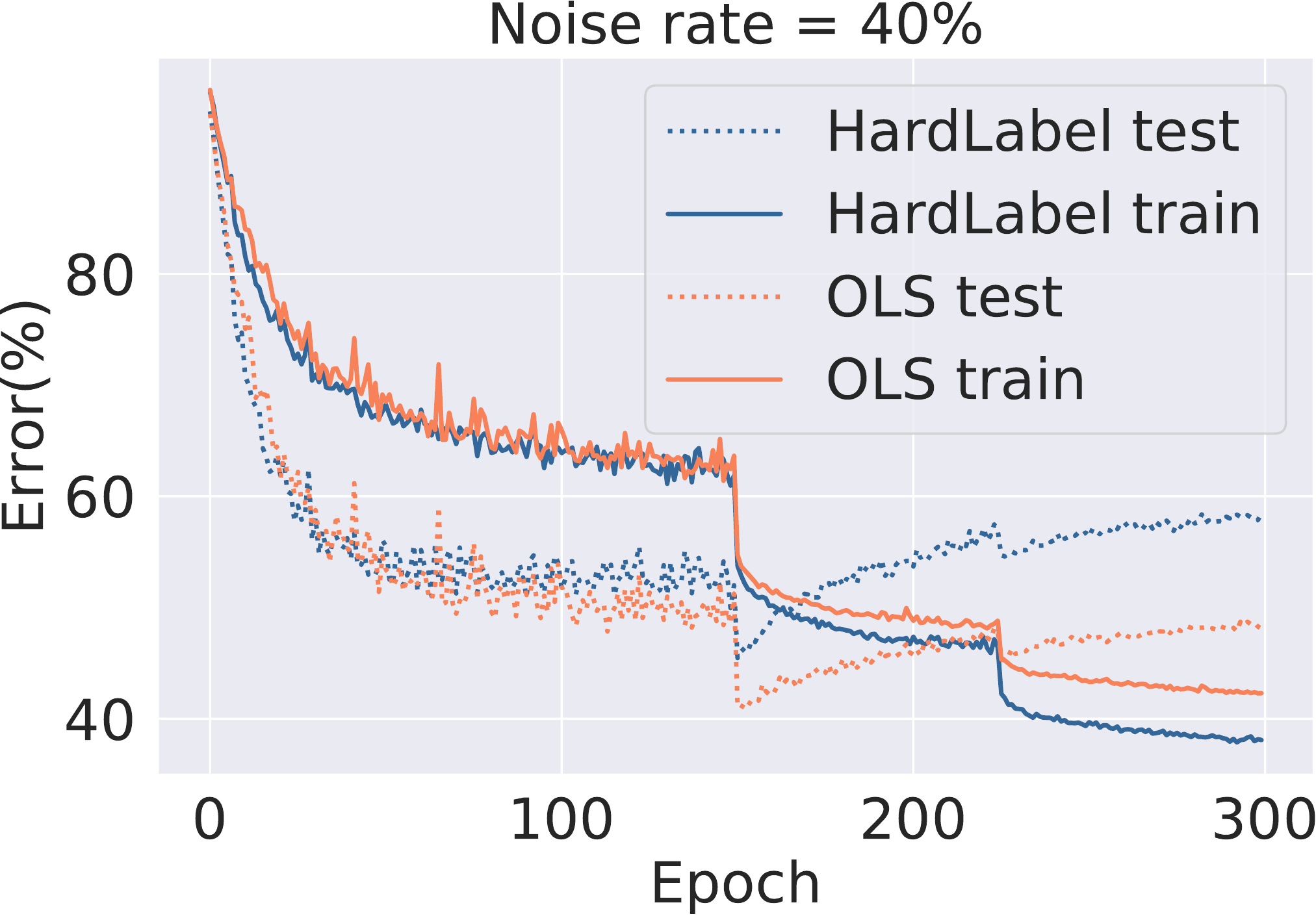}
			\end{overpic}
		}
		\subfigure{ 
			\begin{overpic}[width=0.31\linewidth, tics=5]{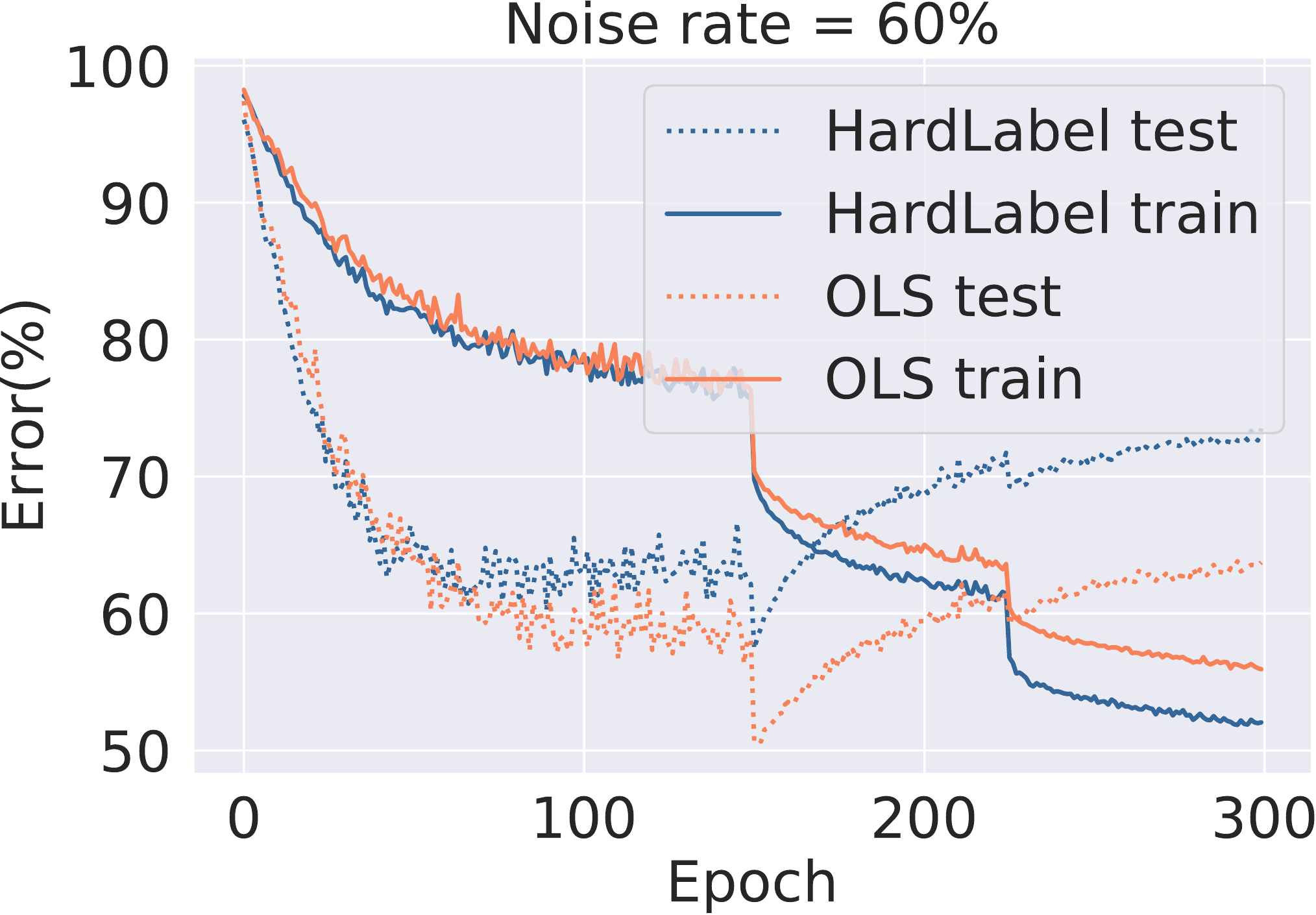}
			\end{overpic}
		}
	\end{small}
	\vskip -0.1in
	\caption{We display the training error and test error under different noise rates (20\%, 40\%, 60\%).} \label{fig:noise}
	\vskip -0.1in
\end{figure*}

In \tabref{tab:noisylabels}, we report the classification 
results based on the ResNet-56 model when the noisy rate is set
to $\{0\%, 20\%, 40\%, 60\%, 80\%\}$, respectively.
It can be seen that our method achieves comparable results 
with those methods~\cite{sl,Ren2018reweight,metaweight,unsupervised} 
that are specifically designed for noisy labels.
Comparing with LS, our method achieves stable improvement 
under different noisy rates.
We also visualize training and test errors during the training 
process.
As shown in \figref{fig:noise}, our method achieves higher 
training errors than models trained with hard labels and LS.
However, our method has lower test errors.
This demonstrates that our method can effectively reduce the 
overfitting to noisy samples.

\begin{figure}[!thp]
	\vskip -0.2in
	\begin{small}
		\begin{center}
			\subfigure{ 
				\begin{overpic}[width=0.6\linewidth, tics=5]{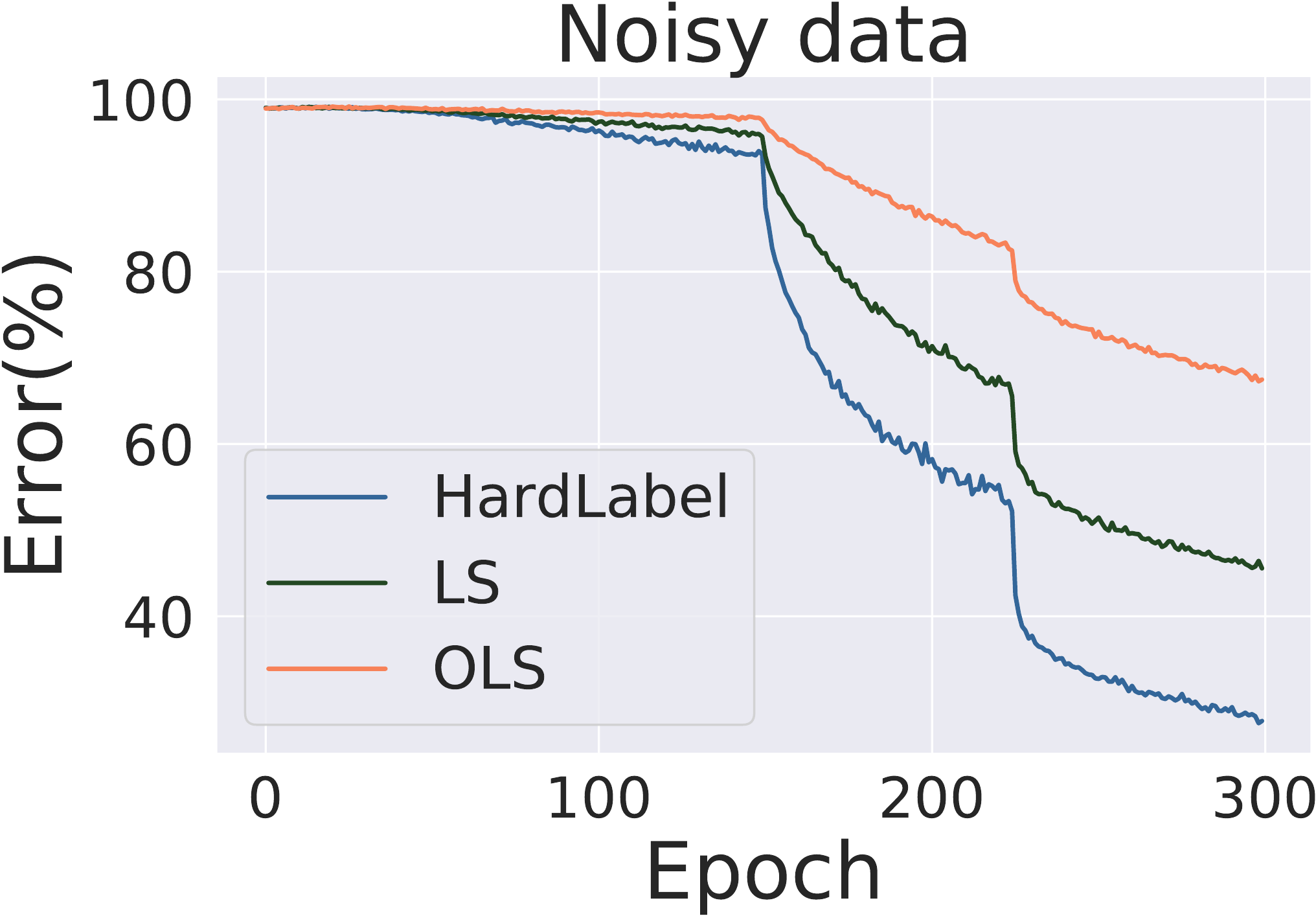}
				\end{overpic}
			}
		\end{center}
		
	\end{small}
	\vskip -0.1in
	\caption{We show the error rate in the training process on all images with wrong labels in the training set.
			The error rate calculation is still based on the wrong labels, \ie the labels of the images are wrong.
			Experiments are conducted on CIFAR-100 under a 40\% noise rate.} \label{fig:whyisnoisy}
	\vskip -0.1in
\end{figure}

Furthermore,
as shown in~\figref{fig:whyisnoisy},
we visualize the Top-1 Error for the set of samples with wrong labels in the training set during the training process.
Note that the error rate calculation uses the wrong labels,
\ie the higher the error rate for the wrong labels, the lower the fit to the wrong labels.
Our method fits the wrong labels worse than baselines.
This phenomenon demonstrates that our method is robust to noisy labels by reducing the fitting to wrong labels.
Our method brings intra-class constraints,
which makes it more difficult for the model to fit the data with the wrong labels.

\subsection{Robustness to Adversarial Attacks} \label{sec:robustness}
\begin{figure*}[!tp] 
\begin{small}
	\begin{overpic}[width=1.0\textwidth,tics=2]{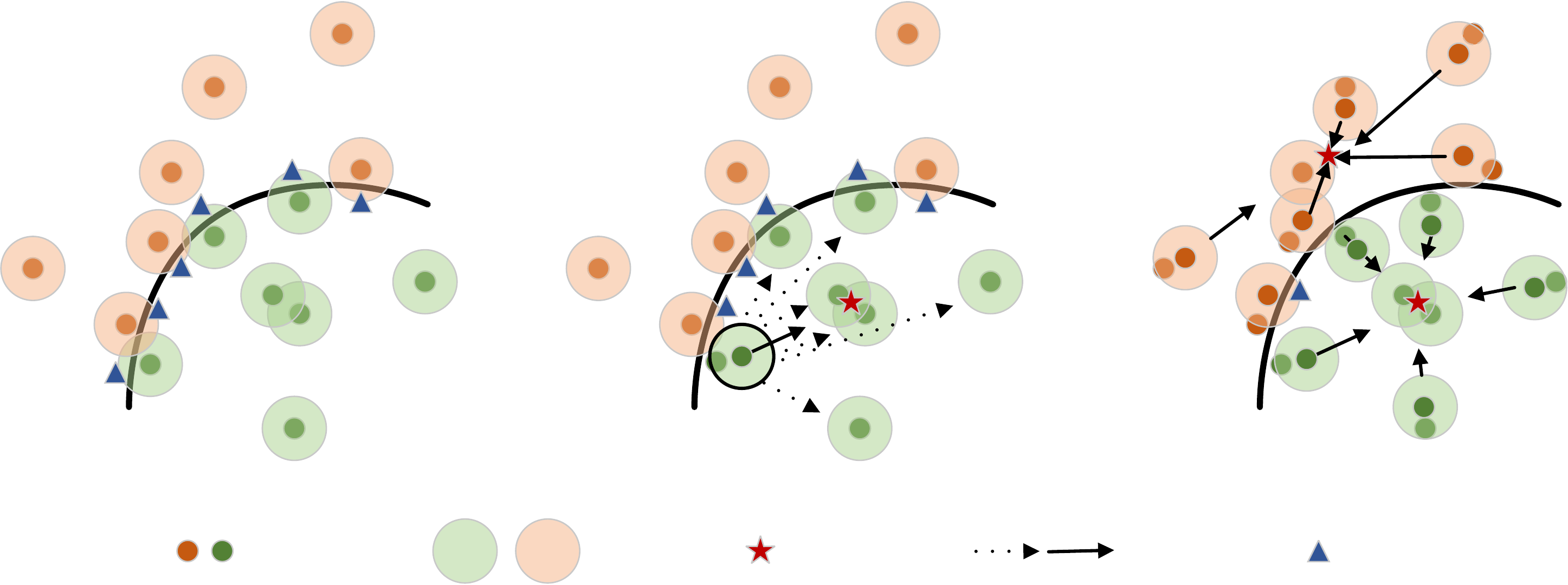}
		\put(1, 6){(a)Training w/o intra-class constraints}
		\put(34, 6){(b)Training w/ intra-class constraints for a sample}
		\put(75, 6){(c)Training w/ intra-class constraints}
		\put(15,1.5){samples}
		\put(37, 2.5){neighborhood}
		\put(37, 0.5){($\epsilon$-ball)}
		\put(50,2.5){optimization}
		\put(50,0.5){target}
		\put(72,2.5){optimization}
		\put(72,0.5){direction}
		\put(85,2.5){adversarial}
		\put(85,0.5){example}
	\end{overpic}
\end{small}
\vskip -0.1in
\caption{The impact of intra-class constraints on training.
	(a) After being trained with hard labels,
	it is easy to get the adversarial example with the $\epsilon$-ball of it crossing the decision boundary.
	(b) In one training iteration, 
	the OLS makes consistency constraints between current training samples and samples in the same class,
	which makes the current training samples far away from the decision boundary.
	The dotted line indicates that each sample in the same class will have a consistency constraint on the current training sample.
	The solid line denotes the optimization direction of this training sample.
	(c) The intra-class constraints make the samples in the same class closer,
	and further away from the decision boundary. It will be more difficult to find adversarial samples.
} \label{fig:adv}
\end{figure*}

In this section, we first explain why our method is more robust to adversarial attacks.
To get the adversarial example for $x$,
FGSM looks for points that cross the decision boundary in the neighborhood $\epsilon$-ball of sample $x$,
so that $x$ is misclassified.
The adversarial example $x_{adv}$ could be denoted as:
\begin{equation}
\begin{aligned}
x_{adv} = x + \gamma sign(\nabla_{x} L(\theta, x, y)),
\label{eq:fgsm}
\end{aligned}
\end{equation}
where $L$ denotes the loss function and $\gamma$ is a coefficient denoting the optimization step.
The $sign()$ is
\begin{equation}
sign(z) = \begin{cases}
1 & \text{ if } z > 0 \\
0 & \text{ if } z = 0 \\
-1 & \text{ if } z < 0.
\end{cases} \label{eq:sign}
\end{equation}
As shown in~\figref{fig:adv}(a),
the purpose of FGSM~\cite{fgsm} is to find a perturbation point that can be misclassified
in the neighborhood ($\epsilon$-ball) for each sample.
Therefore,
it is easy to find the adversarial example for samples near the decision boundary.
In our method,
for each class $k$,
the soft label is accumulated by all predictions of samples in the same class.
The loss function as~\eqref{eq:intraclass} indicates that all correctly classified samples $x_i$ will impose the intra-class
constraints to the current training sample $x_i$.
The constraints encourage the samples belonging to the same class to be much closer.
As shown in~\figref{fig:adv}(b),
in one training iteration,
the intra-class constraints in our method will drive the current training sample
to become more closer with samples in the same class.
Thus,
the intra-class will lead to the compactification of samples in the same class,
as shown in~\figref{fig:adv}(c).
Compared to~\figref{fig:adv}(a),
the number of samples near the decision boundary will be reduced.
This will make the model more robust against adversarial attacks.

And we evaluate the robustness of the models trained by different methods 
against adversarial attack algorithms on CIFAR-10 and ImageNet,
respectively.
We use the Fast Gradient Sign Method (FGSM)~\cite{fgsm} and 
Projected Gradient Descent (PGD)~\cite{pgd} to generate adversarial 
samples.
For FGSM, we keep its default setup. 
Therefore, the $\ell_\infty$ bound is set to 8 for all methods.
For PGD, we apply the same experimental setup as in~\cite{Peterson_2019_ICCV} 
except that we increase the iteration times to 20, which is enough 
to get better attack effects.
\begin{table}[t!]
  \centering
  \renewcommand{\tabcolsep}{0.5mm} 
  \caption{Robustness to adversarial attack on CIFAR-10. 
  	We use FGSM and PGD algorithms to attack ResNet-29 
  	trained on CIFAR-10, respectively. 
    We set the iteration times of PGD attack algorithm as 20.
  }\label{tab:cifar10adversarial}
  \begin{tabular}{lccc}
    \toprule
    Method & \makecell{ResNet-29\\Top-1 Err(\%)} & \makecell{\textbf{+} FGSM\\Top-1 Err(\%)} & \makecell{\textbf{+} PGD\\Top-1 Err(\%)} \\
    \midrule 
    Hard Label & 7.18 & 82.46 & 93.18 \\
    Bootsoft~\cite{bootstrap} & 6.91 & 79.83 & 92.57 \\
    Boothard~\cite{bootstrap} & 7.73 & 82.68 & 90.01 \\
    Symmetric Cross Entropy~\cite{sl} & 8.66 & 77.68 & 93.96 \\
    LS~\cite{christian2016rethinking} & 6.81 & 79.48 & 87.32 \\
    OLS & \textbf{6.46} & \textbf{60.39} & \textbf{76.29} \\
    \bottomrule
  \end{tabular}     
\end{table}
\begin{table}[t!]
  \centering
  \renewcommand{\tabcolsep}{0.8mm} 
	\caption{Top-1 and Top-5 Error(\%) of ResNet-50 on ImageNet after the 
	  adversarial attack.
		For two adversarial attack algorithms, FGSM and PGD,
		we keep their default setting.
		We set the iteration times of PGD attack algorithm as 20.
	}\label{tab:imagenetadversarial}
  \begin{tabular}{lcccc}
		\toprule
		\multirow{2}*{ResNet-50} & \multicolumn{2}{c}{\textbf{+} FGSM} & \multicolumn{2}{c}{\textbf{+} PGD} \\
		\cmidrule(lr{0.25em}){2-3} \cmidrule(lr{0.25em}){4-5} 
		~ & {Top-1 Err(\%)} & {Top-5 Err(\%)} & {Top-1 Err(\%)} & {Top-5 Err(\%)} \\
		\midrule
		Hard Label & 91.07 & 66.21 & 94.93 & 31.82 \\
		Bootsoft~\cite{bootstrap} & 91.29 & 67.29 & 94.56 & 31.07 \\
		LS~\cite{christian2016rethinking} & \textbf{74.44} & 50.63 & 80.31 & 24.46 \\
		OLS & 75.79 & \textbf{48.13} & \textbf{74.43} & \textbf{22.14} \\
		\bottomrule
	\end{tabular}
\end{table}

In \tabref{tab:cifar10adversarial}, we have reported the Top-1 Error 
after the adversarial attack from the FGSM and PGD algorithms 
on the CIFAR-10 dataset.
After the FGSM and PGD attack, the models trained with our method 
keep the lowest Top-1 Error rate.
We can see that the models trained with our OLS algorithm are much more 
robust to the adversarial attack than those trained with other methods.
Moreover, we apply the same experiments on ImageNet, as shown in \tabref{tab:imagenetadversarial}.
Compared with the hard label,
OLS achieves an average 17.9\% gain in terms of Top-1 Error
and an average 13.9\% gain in terms of Top-5 Error.
Our method can also outperform LS~\cite{christian2016rethinking} by
2.3\% and by 2.4\% on Top-1 Error and Top-5 Error,
respectively.
We argue that the soft labels generated in our algorithm
contain similarities between categories,
making the distances of the embedding of samples in the same class
closer.
Experiments show that OLS can effectively improve the robustness of the model to adversarial examples.

\begin{table}[t!]
  \centering
  \renewcommand{\tabcolsep}{3.5mm} 
  \caption{Object detection results. We train YOLO \cite{redmon2016you} on PASCAL VOC dataset. 
  }\label{tab:det}
  \begin{tabular}{lccc}
    \toprule
    Method & Hard Label & LS~\cite{christian2016rethinking} & OLS \\
    \midrule 
    mAP (\%)   & 81.6  & 82.3  & \textbf{82.7}  \\
    \bottomrule
  \end{tabular}     
\end{table}

\subsection{Object Detection} \label{sec:objectdet}
Our OLS can be easily applied to the object detection framework~\cite{CRM2020Fang,FPR2019Sun,faster2017Ren,liu2016ssd,tian2019fcos}.
We select YOLO \cite{redmon2016you} as our basic detector.
We train the detector on the popular PASCAL VOC dataset 
\cite{everingham2010pascal}.
As shown in \tabref{tab:det}, when YOLO is equipped 
with our OLS, it obtains a 1.1\% gain over the hard label
and a 0.4\% gain over LS in terms of mean 
average precision (mAP),
indicating OLS has stronger regularization ability than LS
on the object detection.

\textbf{Implement details.}
We use MobileNetv2~\cite{sandler18mobile} as the backbone of YOLO~\cite{redmon2016you}.
We regard the combination of the training set and validation set from PASCAL VOC 2012 
and PASCAL VOC 2007 as the training set.
And we test the model on the PASCAL VOC 2007 test set.
During training, we use standard training strategies,
including warming up, multi-scale training, random crop, etc.
We train the model for 120 epochs using SGD optimizer with an initial learning rate 0.0001
and cosine learning rate decay schedule.
During tests,
we also use multi-scale inference.

\begin{table*}[!tp]
  \centering
  \renewcommand{\tabcolsep}{4.6mm}
  \caption{Top-1 Error(\%) and Expected Calibration Error(ECE) on CIFAR-100. 
    Lower is better. 
  }
  \begin{tabular}{lcccccc}
    \toprule
    \multirow{2}*{Method} & \multicolumn{2}{c}{ResNet-56} & \multicolumn{2}{c}{ResNet-74} & \multicolumn{2}{c}{ResNet-110}\\
    \cmidrule(lr{0.25em}){2-3} \cmidrule(lr{0.25em}){4-5} 
    \cmidrule(lr{0.25em}){6-7} ~ & {Top-1 Error(\%)} & 
    {ECE} & {Top-1 Error(\%)} & {ECE} & {Top-1 Error(\%)} & {ECE} \\
    \midrule
    {Hard Label} 
    & $26.81\pm 0.36$ & $11.37\pm 0.53$
    & $25.86\pm 0.19$ & $12.70\pm 0.76$
    & $25.54\pm 0.44$ & $13.14\pm 1.16$ \\
    {LS~\cite{christian2016rethinking}}
    & $26.37\pm 0.41$ & $3.35\pm 0.86$ 
    & $25.90\pm 0.31$ & $2.37\pm 0.94$ 
    & $25.14\pm 0.31$ & $2.32\pm 1.03$ \\
    {OLS}
    & \bm{$25.24\pm 0.18$} & \bm{$2.85\pm 1.44$}
    & \bm{$24.89\pm 0.08$} & \bm{$1.81\pm 0.85$}
    & \bm{$23.86\pm 0.27$} & \bm{$2.05\pm 0.68$} \\
    \bottomrule
  \end{tabular}
  \label{tab:calibration}
\end{table*}

\subsection{Ablation Study} \label{sec:ablation}
In this subsection, we first conduct experiments to study 
the hyper-parameters in our method.
Then we analyze the relationships among categories 
indicated by our soft labels.
Besides, we also present a variant of OLS.
Finally, we present the calibration effect of our method.
All the experiments are conducted on the CIFAR dataset.

\begin{figure}[!ht]
	\centering
	\begin{small} 
		\centering
		\subfigure[Impact of $\alpha$]{
			\includegraphics[width=0.45\linewidth]{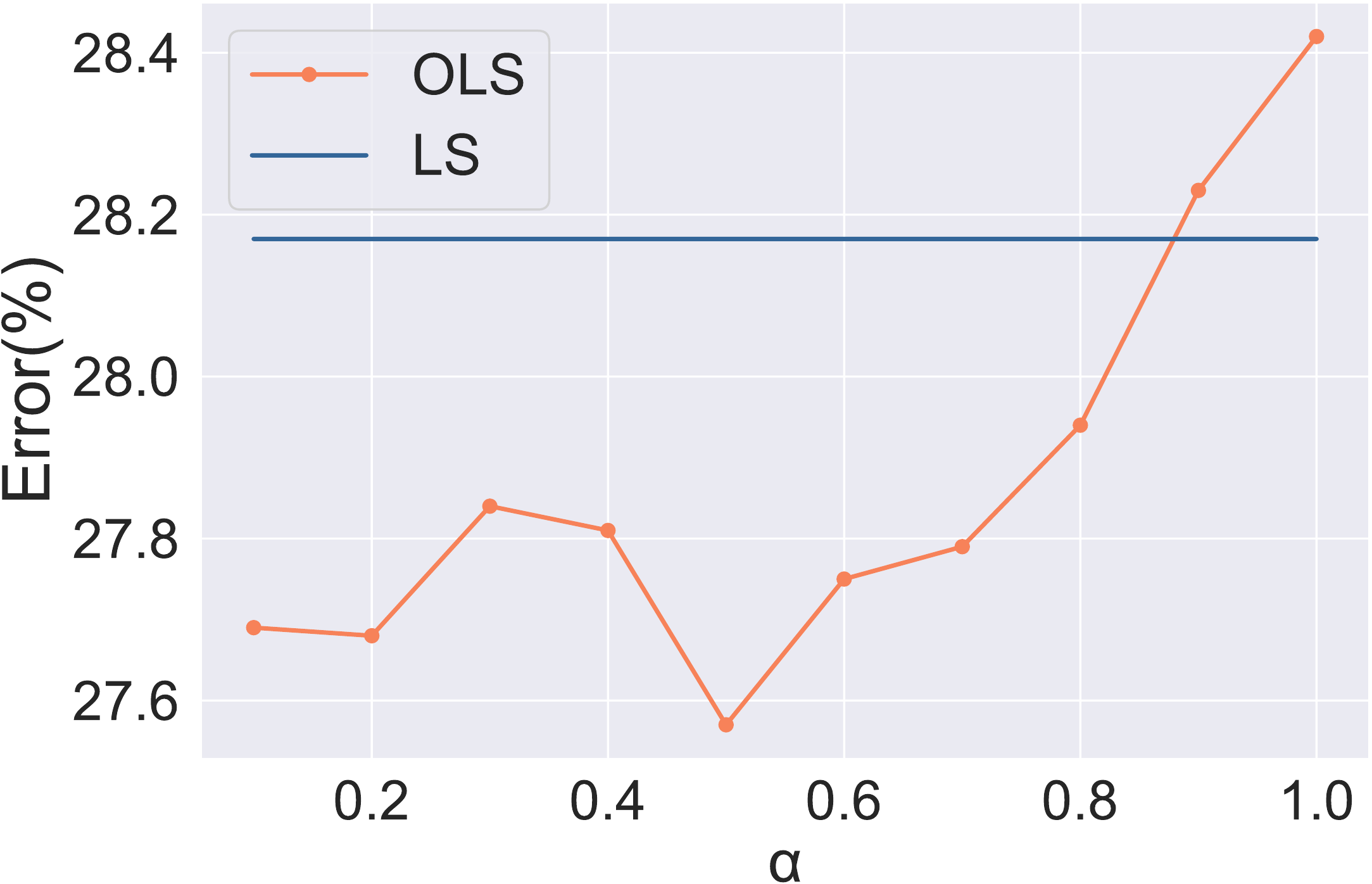}
		}
		\subfigure[Impact of the updating period]{
			\includegraphics[width=0.45\linewidth]{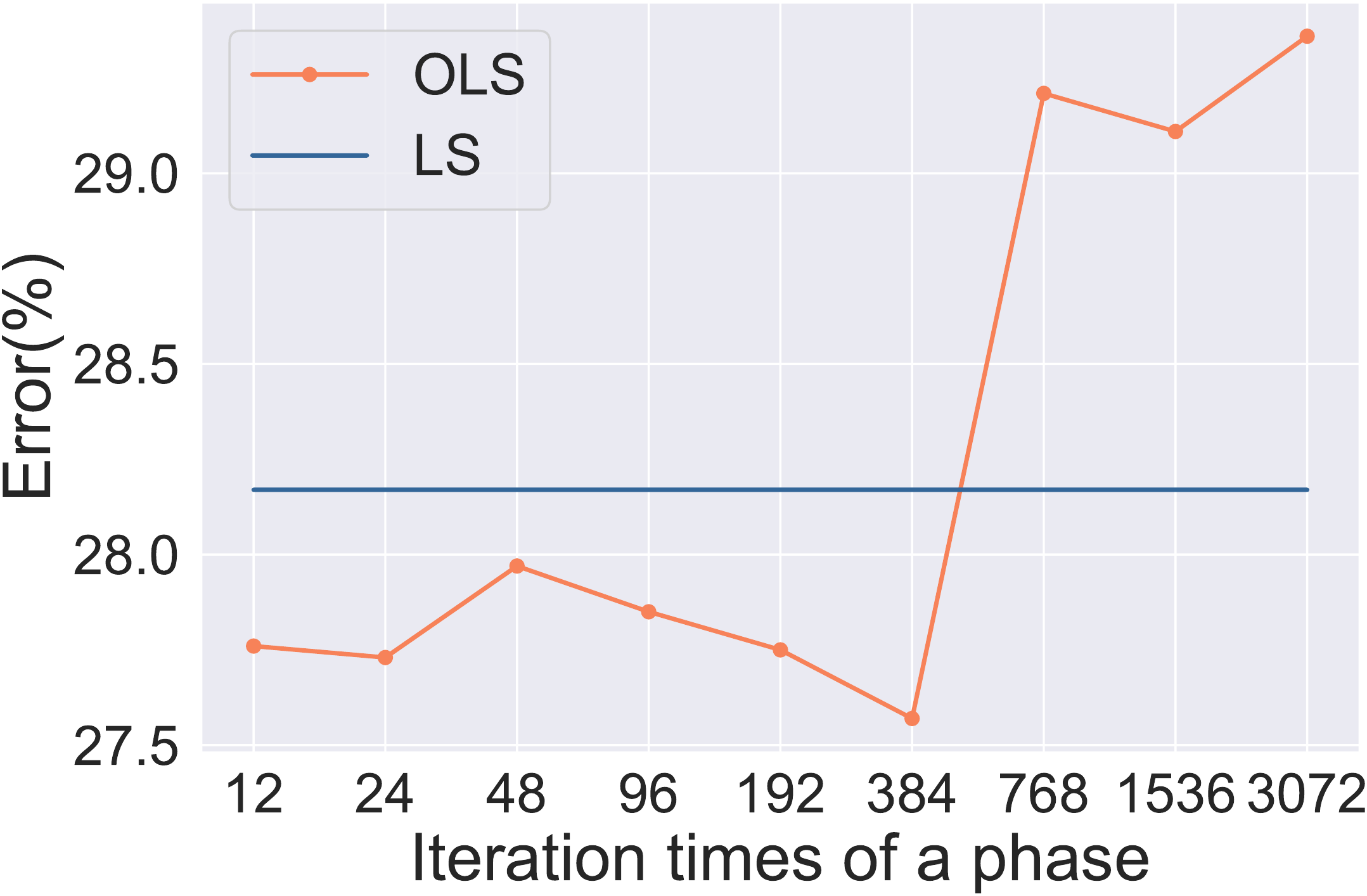}
		}
	\end{small}
	\vskip -0.1in
	\caption{Impact of hyper-parameters. The Top-1 Error of  different $\alpha$ and updating period.
	}
	\label{fig:impactofhyper}
	\vskip -0.1in
\end{figure}

\textbf{Impact of Hyper-parameters.}
We first analyze the hyper-parameter $\alpha$ 
in \eqref{eq:olsloss} using ResNet-29.
Unlike previous experiments that directly set $\alpha$ to 0.5, 
we enumerate possible values with 
$\alpha\in\{0.1, 0.2,\cdots ,1.0\}$.
We plot the experiment results as shown in 
\figref{fig:impactofhyper}(a).
It can be seen that the model achieves the lowest top-1 
error when $\alpha$ is set to 0.5.
Since the model lacks guidelines for the correct category,
we observe that when $\alpha$ is set to 0, the model is 
hard to convergent.
When $\alpha$ changes from 0.1 to 0.5, the error rate gradually 
decreases. 
This fact suggests that the model still needs the correct 
category information provided by the original hard labels.

Moreover, we also conduct experiments to study the impact 
of the updating period for the soft label matrix $S$ 
in the training process.
In the previous experiments in \secref{sec:generalimagerecognition},
we set the updating period to one epoch.
As shown in \figref{fig:impactofhyper}(b), we evaluate 
our approach with different updating periods (
iteration times 
$\in\{12, 24, 48, \cdots , 1536, 3072\}$).
The best performance is obtained when the updating 
period is set to one epoch.
We observe that the classification performance is very close 
when the updating period is less than one epoch
(1 epoch is approximately 384 iterations).
However, when the updating period is longer than one training epoch, 
the performance decreases sharply.
We analyze that with the training of the network, the predictions 
become better and better.
When using more iterations to update soft labels, the relationships 
indicated by the early predictions will be very different from 
that of late ones.
The early predictions become out of date for current training.

\textbf{Importance of relationships among categories.}
We argue that classification models can benefit from soft labels 
that contain the knowledge of relationships among 
different categories.  
Specifically, we utilize a human uncertainty dataset \cite{Peterson_2019_ICCV}
called CIFAR-10H to verify the reliability of the 
relationships among different categories.
CIFAR-10H captures the full distribution of the
labels by collecting votes from more than 50 people 
for each sample in the CIFAR-10 test set.
The human uncertainty labels can be regarded as a kind of
soft label that considers the similarities among different
categories.
They find that models trained on the human uncertainty 
labels will have better accuracy and generalization 
than those trained on hard labels.
To explore the rationality of relationships among categories 
found by our approach, we use KL divergence 
to measure the difference between the predicted probability 
distribution of the model and the human uncertainty distribution 
on CIFAR-10H.

\begin{table}[htp!]
	\centering
	\setlength{\tabcolsep}{2.8mm}
	\caption{
		Multiple evaluation results of the model.
		We first train ResNet-29 with different methods on CIFAR-10.
		We use the average KL divergence to measure 
		the difference between the prediction distribution of the models 
		and human uncertainty on CIFAR-10H test set.
	}
	\begin{tabular}{lcc}
		\toprule
		Method & \makecell{CIFAR-10\\Top-1 Err(\%)}  & 
		\makecell{CIFAR-10H\\KL Divergence} \\
		\midrule 
		Hard Label & 7.18   & 0.2974 \\
		Bootsoft~\cite{bootstrap} & 6.91  & 0.3247  \\
		Boothard~\cite{bootstrap} & 7.73  & 0.3188 \\
		Symmetric Cross Entropy~\cite{sl} & 8.66    & 0.5563  \\
		LS~\cite{christian2016rethinking} & 6.81   & 0.1866  \\
		OLS & \textbf{6.46} & \textbf{0.1399} \\
		\bottomrule
	\end{tabular} \label{tab:similarity}
\end{table} 	

For a fair comparison, we only consider the correctly predicted 
samples by each model, when computing the KL Divergence on CIFAR-10H.
As shown in Table \ref{tab:similarity}, we list the average 
KL divergence of different methods on CIFAR-10H~\cite{Peterson_2019_ICCV}
and Top-1 Error(\%) on CIFAR-10. 
The results show that the prediction distribution of the model 
trained by our method is closer to that of humans.
Also, this indicates that the model trained by our approach 
finds more reasonable and correct relationships among
categories.

\textbf{Sample-level soft labels.}
To verify the effectiveness of the statistical characteristics of 
accumulating model predictions, we use the predicted distribution 
of a single sample (denoted as OLS-Single for short) to regularize 
the training process.
To be specific, for each training sample, we randomly select 
another training sample with the same category. 
We then acquire the randomly selected training sample's predictive 
distribution and utilize this distribution as the soft label 
to serve as supervision for the current training sample.
Based on the ResNet-56,
OLS ($25.24\pm 0.18$) outperforms OLS-Single ($26.18\pm 0.30$)
by about 1\%. 
This result demonstrates that the accumulation of the predictions from 
different samples can well explore the relationships among categories.

\textbf{Calibration effect.}
The confidence calibration is proposed in~\cite{guo2017calibration},
which is used to measure the degree of
overfitting of the model
to the training set.
We use the Expected Calibration Error (ECE)~\cite{guo2017calibration} to measure
the calibration ability of OLS.
In \tabref{tab:calibration}, we report the Top-1 Error(\%) and ECE
on several models,
which denotes our method can calibrate neural networks.
Experimental results show that our method
achieves a lower Top-1 Error than LS 
by an average of 1.14\%.
Meanwhile, our method also achieves lower 
ECE values on three different depth models.
This indicates that the proposed method 
can more effectively prevent over-confident predictions 
and show better calibration capability.

\section{Conclusion} \label{sec:Conclusion}
In this paper, we propose an online label smoothing 
method. 
We utilize the statistics of the intermediate model 
predictions to generate soft labels, which are 
subsequently used to supervise the model.
Our soft labels considering the relationships among 
categories are effective in preventing the overfitting problem 
of DNNs to the training set.
We evaluate our OLS on CIFAR, ImageNet and four fine-grained datasets, 
respectively.
On CIFAR-100, ResNeXt-2x64d trained with our OLS achieves 
18.81\% Top-1 Error, which brings an 2.11\% 
performance gain. 
On ImageNet dataset, our OLS brings 1.4\% and 1.02\% 
performance gains to ResNet-50 and ResNet-101, respectively.
On four fine-grained datasets,
OLS outperforms the hard label by 2\% in terms of Top-1 Error.
%


\bibliographystyle{IEEEtran}
\bibliography{OLS}

\begin{thebibliography}{10}
\providecommand{\url}[1]{#1}
\csname url@samestyle\endcsname
\providecommand{\newblock}{\relax}
\providecommand{\bibinfo}[2]{#2}
\providecommand{\BIBentrySTDinterwordspacing}{\spaceskip=0pt\relax}
\providecommand{\BIBentryALTinterwordstretchfactor}{4}
\providecommand{\BIBentryALTinterwordspacing}{\spaceskip=\fontdimen2\font plus
\BIBentryALTinterwordstretchfactor\fontdimen3\font minus
  \fontdimen4\font\relax}
\providecommand{\BIBforeignlanguage}[2]{{%
\expandafter\ifx\csname l@#1\endcsname\relax
\typeout{** WARNING: IEEEtran.bst: No hyphenation pattern has been}%
\typeout{** loaded for the language `#1'. Using the pattern for}%
\typeout{** the default language instead.}%
\else
\language=\csname l@#1\endcsname
\fi
#2}}
\providecommand{\BIBdecl}{\relax}
\BIBdecl

\bibitem{vgg}
K.~Simonyan and A.~Zisserman, ``Very deep convolutional networks for
  large-scale image recognition,'' in \emph{Int. Conf. Learn. Represent.},
  2015.

\bibitem{he2016deep}
K.~He, X.~Zhang, S.~Ren, and J.~Sun, ``Deep residual learning for image
  recognition,'' in \emph{IEEE Conf. Comput. Vis. Pattern Recog.}, 2016, pp.
  770--778.

\bibitem{huang2017densely}
G.~Huang, Z.~Liu, L.~Van Der~Maaten, and K.~Q. Weinberger, ``Densely connected
  convolutional networks,'' in \emph{IEEE Conf. Comput. Vis. Pattern Recog.},
  2017, pp. 4700--4708.

\bibitem{xie2017aggregated}
S.~Xie, R.~Girshick, P.~Doll{\'a}r, Z.~Tu, and K.~He, ``Aggregated residual
  transformations for deep neural networks,'' in \emph{IEEE Conf. Comput. Vis.
  Pattern Recog.}, 2017, pp. 1492--1500.

\bibitem{hu2018squeeze}
J.~Hu, L.~Shen, and G.~Sun, ``Squeeze-and-excitation networks,'' in \emph{IEEE
  Conf. Comput. Vis. Pattern Recog.}, 2018, pp. 7132--7141.

\bibitem{sandler18mobile}
M.~Sandler, A.~G. Howard, M.~Zhu, A.~Zhmoginov, and L.~Chen, ``Mobilenetv2:
  Inverted residuals and linear bottlenecks,'' in \emph{IEEE Conf. Comput. Vis.
  Pattern Recog.}, 2018, pp. 4510--4520.

\bibitem{gao2019res2net}
S.-H. Gao, M.-M. Cheng, K.~Zhao, X.-Y. Zhang, M.-H. Yang, and P.~Torr,
  ``Res2net: A new multi-scale backbone architecture,'' \emph{IEEE Trans.
  Pattern Anal. Mach. Intell.}, vol.~43, no.~2, pp. 652--662, 2020.

\bibitem{krizhevsky2009learning}
A.~Krizhevsky and G.~Hinton, ``Learning multiple layers of features from tiny
  images,'' University of Toronto, Toronto, Ontario, Tech. Rep.~0, 2009.

\bibitem{deng2009imagenet}
J.~Deng, W.~Dong, R.~Socher, L.-J. Li, K.~Li, and L.~Fei-Fei, ``Imagenet: A
  large-scale hierarchical image database,'' in \emph{IEEE Conf. Comput. Vis.
  Pattern Recog.}, 2009, pp. 248--255.

\bibitem{christian2016rethinking}
C.~Szegedy, V.~Vanhoucke, S.~Ioffe, J.~Shlens, and Z.~Wojna, ``Rethinking the
  inception architecture for computer vision,'' in \emph{IEEE Conf. Comput.
  Vis. Pattern Recog.}, 2016, pp. 2818--2826.

\bibitem{bootstrap}
S.~E. Reed, H.~Lee, D.~Anguelov, C.~Szegedy, D.~Erhan, and A.~Rabinovich,
  ``Training deep neural networks on noisy labels with bootstrapping,'' in
  \emph{Int. Conf. Learn. Represent. Worksh.}, 2015.

\bibitem{devries2017cutout}
T.~DeVries and G.~W. Taylor, ``Improved regularization of convolutional neural
  networks with cutout,'' \emph{arXiv preprint arXiv:1708.04552}, 2017.

\bibitem{zhang2018mixup}
H.~Zhang, M.~Cisse, Y.~N. Dauphin, and D.~Lopez-Paz, ``mixup: Beyond empirical
  risk minimization,'' in \emph{Int. Conf. Learn. Represent.}, 2018.

\bibitem{ghiasi2018dropblock}
G.~Ghiasi, T.-Y. Lin, and Q.~V. Le, ``Dropblock: A regularization method for
  convolutional networks,'' in \emph{Adv. Neural Inform. Process. Syst.}, 2018,
  pp. 10\,727--10\,737.

\bibitem{yamada2019shakedrop}
Y.~Yamada, M.~Iwamura, T.~Akiba, and K.~Kise, ``Shakedrop regularization for
  deep residual learning,'' \emph{IEEE Access}, pp. 186\,126--186\,136, 2019.

\bibitem{qi2020loss}
G.-J. Qi, ``Loss-sensitive generative adversarial networks on lipschitz
  densities,'' \emph{Int. J. Comput. Vis.}, vol. 128, no.~5, pp. 1118--1140,
  2020.

\bibitem{hinton15KD}
G.~Hinton, O.~Vinyals, and J.~Dean, ``Distilling the knowledge in a neural
  network,'' in \emph{Adv. Neural Inform. Process. Syst. Worksh.}, 2015.

\bibitem{nilsback2008automated}
M.-E. Nilsback and A.~Zisserman, ``Automated flower classification over a large
  number of classes,'' in \emph{2008 Sixth Indian Conference on Computer
  Vision, Graphics \& Image Processing}, 2008, pp. 722--729.

\bibitem{WahCUB_200_2011}
C.~Wah, S.~Branson, P.~Welinder, P.~Perona, and S.~Belongie, ``{The
  Caltech-UCSD Birds-200-2011 Dataset},'' California Institute of Technology,
  Tech. Rep. CNS-TR-2011-001, 2011.

\bibitem{krause20133d}
J.~Krause, M.~Stark, J.~Deng, and L.~Fei-Fei, ``3d object representations for
  fine-grained categorization,'' in \emph{Int. Conf. Comput. Vis. Worksh.},
  2013, pp. 554--561.

\bibitem{maji2013fine}
S.~Maji, J.~Kannala, E.~Rahtu, M.~Blaschko, and A.~Vedaldi, ``A database for
  fine-grained aircraft recognition,'' in \emph{IEEE Conf. Comput. Vis. Pattern
  Recog. Worksh.}, June 2013.

\bibitem{efficientNet}
M.~Tan and Q.~V. Le, ``Efficientnet: Rethinking model scaling for convolutional
  neural networks,'' in \emph{Int. Conf. Mech. Learn.}, vol.~97, 2019, pp.
  6105--6114.

\bibitem{zhao2020exploring}
H.~Zhao, J.~Jia, and V.~Koltun, ``Exploring self-attention for image
  recognition,'' in \emph{IEEE Conf. Comput. Vis. Pattern Recog.}, 2020, pp.
  10\,076--10\,085.

\bibitem{zhang2019byot}
L.~Zhang, J.~Song, A.~Gao, J.~Chen, C.~Bao, and K.~Ma, ``Be your own teacher:
  Improve the performance of convolutional neural networks via self
  distillation,'' in \emph{Int. Conf. Comput. Vis.}, 2019, pp. 3712--3721.

\bibitem{xu2019data}
T.-B. Xu and C.-L. Liu, ``Data-distortion guided self-distillation for deep
  neural networks,'' in \emph{AAAI Conf. Artif. Intell.}, 2019, pp. 5565--5572.

\bibitem{Xie2016disturb}
L.~Xie, J.~Wang, Z.~Wei, M.~Wang, and Q.~Tian, ``Disturblabel: Regularizing cnn
  on the loss layer,'' in \emph{IEEE Conf. Comput. Vis. Pattern Recog.}, 2016,
  pp. 4753--4762.

\bibitem{dubey2018pairwise}
A.~Dubey, O.~Gupta, P.~Guo, R.~Raskar, R.~Farrell, and N.~Naik, ``Pairwise
  confusion for fine-grained visual classification,'' in \emph{Eur. Conf.
  Comput. Vis.}, 2018, pp. 70--86.

\bibitem{li2020reconstruction}
C.~{Li}, C.~{Liu}, L.~{Duan}, P.~{Gao}, and K.~{Zheng}, ``Reconstruction
  regularized deep metric learning for multi-label image classification,''
  \emph{IEEE Trans. Neural Netw. Learn Syst.}, vol.~31, no.~7, pp. 2294--2303,
  2020.

\bibitem{zhang2019aet}
L.~Zhang, G.-J. Qi, L.~Wang, and J.~Luo, ``Aet vs. aed: Unsupervised
  representation learning by auto-encoding transformations rather than data,''
  in \emph{IEEE Conf. Comput. Vis. Pattern Recog.}, 2019, pp. 2547--2555.

\bibitem{qi2020learning}
G.-J. Qi, L.~Zhang, F.~Lin, and X.~Wang, ``Learning generalized transformation
  equivariant representations via autoencoding transformations,'' \emph{IEEE
  Trans. Pattern Anal. Mach. Intell.}, 2020.

\bibitem{qi2019avt}
G.-J. Qi, L.~Zhang, C.~W. Chen, and Q.~Tian, ``Avt: Unsupervised learning of
  transformation equivariant representations by autoencoding variational
  transformations,'' in \emph{Int. Conf. Comput. Vis.}, 2019, pp. 8130--8139.

\bibitem{wang2020enaet}
X.~Wang, D.~Kihara, J.~Luo, and G.-J. Qi, ``Enaet: A self-trained framework for
  semi-supervised and supervised learning with ensemble transformations,''
  \emph{IEEE Trans. Image Process.}, 2020.

\bibitem{passalis19unsupervised}
N.~{Passalis} and A.~{Tefas}, ``Unsupervised knowledge transfer using
  similarity embeddings,'' \emph{IEEE Trans. Neural Netw. Learn Syst.},
  vol.~30, no.~3, pp. 946--950, 2019.

\bibitem{tommaso2018ban}
T.~Furlanello, Z.~C. Lipton, M.~Tschannen, L.~Itti, and A.~Anandkumar,
  ``Born-again neural networks,'' in \emph{Int. Conf. Mech. Learn.}, 2018, pp.
  1602--1611.

\bibitem{ge2020distilling}
S.~{Ge}, Z.~{Luo}, C.~{Zhang}, Y.~{Hua}, and D.~{Tao}, ``Distilling channels
  for efficient deep tracking,'' \emph{IEEE Trans. Image Process.}, vol.~29,
  pp. 2610--2621, 2020.

\bibitem{wang2020real}
N.~{Wang}, W.~{Zhou}, Y.~{Song}, C.~{Ma}, and H.~{Li}, ``Real-time correlation
  tracking via joint model compression and transfer,'' \emph{IEEE Trans. Image
  Process.}, vol.~29, pp. 6123--6135, 2020.

\bibitem{ge2019low}
S.~{Ge}, S.~{Zhao}, C.~{Li}, and J.~{Li}, ``Low-resolution face recognition in
  the wild via selective knowledge distillation,'' \emph{IEEE Trans. Image
  Process.}, vol.~28, no.~4, pp. 2051--2062, 2019.

\bibitem{peng2019few}
Z.~Peng, Z.~Li, J.~Zhang, Y.~Li, G.-J. Qi, and J.~Tang, ``Few-shot image
  recognition with knowledge transfer,'' in \emph{Int. Conf. Comput. Vis.},
  2019, pp. 441--449.

\bibitem{yao2019deep}
J.~{Yao}, J.~{Wang}, I.~W. {Tsang}, Y.~{Zhang}, J.~{Sun}, C.~{Zhang}, and
  R.~{Zhang}, ``Deep learning from noisy image labels with quality embedding,''
  \emph{IEEE Trans. Image Process.}, vol.~28, no.~4, pp. 1909--1922, 2019.

\bibitem{duncan1992reinforcement}
J.~S. {Duncan} and T.~{Birkholzer}, ``Reinforcement of linear structure using
  parametrized relaxation labeling,'' \emph{IEEE Trans. Pattern Anal. Mach.
  Intell.}, vol.~14, no.~5, pp. 502--515, 1992.

\bibitem{wang2018multiclass}
R.~{Wang}, T.~{Liu}, and D.~{Tao}, ``Multiclass learning with partially
  corrupted labels,'' \emph{IEEE Trans. Neural Netw. Learn Syst.}, vol.~29,
  no.~6, pp. 2568--2580, 2018.

\bibitem{wei2020harness}
Y.~{Wei}, C.~{Gong}, S.~{Chen}, T.~{Liu}, J.~{Yang}, and D.~{Tao}, ``Harnessing
  side information for classification under label noise,'' \emph{IEEE Trans.
  Neural Netw. Learn Syst.}, vol.~31, no.~9, pp. 3178--3192, 2020.

\bibitem{han2018progressive}
B.~{Han}, I.~W. {Tsang}, L.~{Chen}, C.~P. {Yu}, and S.~{Fung}, ``Progressive
  stochastic learning for noisy labels,'' \emph{IEEE Trans. Neural Netw. Learn
  Syst.}, vol.~29, no.~10, pp. 5136--5148, 2018.

\bibitem{tanaka2018joint}
D.~{Tanaka}, D.~{Ikami}, T.~{Yamasaki}, and K.~{Aizawa}, ``Joint optimization
  framework for learning with noisy labels,'' in \emph{IEEE Conf. Comput. Vis.
  Pattern Recog.}, 2018, pp. 5552--5560.

\bibitem{han2019deep}
J.~Han, P.~Luo, and X.~Wang, ``Deep self-learning from noisy labels,'' in
  \emph{Int. Conf. Comput. Vis.}, 2019, pp. 5138--5147.

\bibitem{Ren2018reweight}
M.~Ren, W.~Zeng, B.~Yang, and R.~Urtasun, ``Learning to reweight examples for
  robust deep learning,'' in \emph{Int. Conf. Mech. Learn.}, 2018, pp.
  4334--4343.

\bibitem{metaweight}
J.~Shu, Q.~Xie, L.~Yi, Q.~Zhao, S.~Zhou, Z.~Xu, and D.~Meng, ``Meta-weight-net:
  Learning an explicit mapping for sample weighting,'' in \emph{Adv. Neural
  Inform. Process. Syst.}, 2019, pp. 1919--1930.

\bibitem{liu2015classification}
T.~Liu and D.~Tao, ``Classification with noisy labels by importance
  reweighting,'' \emph{IEEE Trans. Pattern Anal. Mach. Intell.}, vol.~38,
  no.~3, pp. 447--461, 2015.

\bibitem{sl}
Y.~Wang, X.~Ma, Z.~Chen, Y.~Luo, J.~Yi, and J.~Bailey, ``Symmetric cross
  entropy for robust learning with noisy labels,'' in \emph{Int. Conf. Comput.
  Vis.}, 2019, pp. 322--330.

\bibitem{tanno2019learn}
R.~{Tanno}, A.~{Saeedi}, S.~{Sankaranarayanan}, D.~C. {Alexander}, and
  N.~{Silberman}, ``Learning from noisy labels by regularized estimation of
  annotator confusion,'' in \emph{IEEE Conf. Comput. Vis. Pattern Recog.},
  2019, pp. 11\,236--11\,245.

\bibitem{unsupervised}
E.~Arazo, D.~Ortego, P.~Albert, N.~O'Connor, and K.~Mcguinness, ``Unsupervised
  label noise modeling and loss correction,'' in \emph{Int. Conf. Mech.
  Learn.}, 2019, pp. 312--321.

\bibitem{zhang2018improving}
J.~{Zhang}, V.~S. {Sheng}, T.~{Li}, and X.~{Wu}, ``Improving crowdsourced label
  quality using noise correction,'' \emph{IEEE Trans. Neural Netw. Learn
  Syst.}, vol.~29, no.~5, pp. 1675--1688, 2018.

\bibitem{fang2018data}
M.~{Fang}, T.~{Zhou}, J.~{Yin}, Y.~{Wang}, and D.~{Tao}, ``Data subset
  selection with imperfect multiple labels,'' \emph{IEEE Trans. Neural Netw.
  Learn Syst.}, vol.~30, no.~7, pp. 2212--2221, 2019.

\bibitem{yi2019probabilistic}
K.~Yi and J.~Wu, ``Probabilistic end-to-end noise correction for learning with
  noisy labels,'' in \emph{IEEE Conf. Comput. Vis. Pattern Recog.}, 2019, pp.
  7017--7025.

\bibitem{MullerKH19}
R.~M{\"{u}}ller, S.~Kornblith, and G.~E. Hinton, ``When does label smoothing
  help?'' in \emph{Adv. Neural Inform. Process. Syst.}, 2019, pp. 4696--4705.

\bibitem{tsne}
L.~v.~d. Maaten and G.~Hinton, ``Visualizing data using t-sne,'' \emph{Journal
  of machine learning research}, pp. 2579--2605, 2008.

\bibitem{yuan2020revisiting}
L.~Yuan, F.~E. Tay, G.~Li, T.~Wang, and J.~Feng, ``Revisiting knowledge
  distillation via label smoothing regularization,'' in \emph{IEEE Conf.
  Comput. Vis. Pattern Recog.}, 2020, pp. 3903--3911.

\bibitem{paszke2017automatic}
A.~Paszke, S.~Gross, S.~Chintala, G.~Chanan, E.~Yang, Z.~DeVito, Z.~Lin,
  A.~Desmaison, L.~Antiga, and A.~Lerer, ``Automatic differentiation in
  pytorch,'' in \emph{Adv. Neural Inform. Process. Syst. Worksh.}, 2017.

\bibitem{hu2020jittor}
S.-M. Hu, D.~Liang, G.-Y. Yang, G.-W. Yang, and W.-Y. Zhou, ``Jittor: a novel
  deep learning framework with meta-operators and unified graph execution,''
  \emph{Science China Information Sciences}, vol.~63, no.~12, pp. 1--21, 2020.

\bibitem{iscen2015comp}
A.~{Iscen}, G.~{Tolias}, P.~{Gosselin}, and H.~{Jégou}, ``A comparison of
  dense region detectors for image search and fine-grained classification,''
  \emph{IEEE Trans. Image Process.}, vol.~24, no.~8, pp. 2369--2381, 2015.

\bibitem{zhang2017fine}
C.~{Zhang}, C.~{Liang}, L.~{Li}, J.~{Liu}, Q.~{Huang}, and Q.~{Tian},
  ``Fine-grained image classification via low-rank sparse coding with general
  and class-specific codebooks,'' \emph{IEEE Trans. Neural Netw. Learn Syst.},
  vol.~28, no.~7, pp. 1550--1559, 2017.

\bibitem{zhang2016weak}
Y.~{Zhang}, X.~{Wei}, J.~{Wu}, J.~{Cai}, J.~{Lu}, V.~{Nguyen}, and M.~N. {Do},
  ``Weakly supervised fine-grained categorization with part-based image
  representation,'' \emph{IEEE Trans. Image Process.}, vol.~25, no.~4, pp.
  1713--1725, 2016.

\bibitem{shi2019fine}
W.~{Shi}, Y.~{Gong}, X.~{Tao}, D.~{Cheng}, and N.~{Zheng}, ``Fine-grained image
  classification using modified dcnns trained by cascaded softmax and
  generalized large-margin losses,'' \emph{IEEE Trans. Neural Netw. Learn
  Syst.}, vol.~30, no.~3, pp. 683--694, 2019.

\bibitem{shu2016image}
X.~Shu, J.~Tang, G.-J. Qi, Z.~Li, Y.-G. Jiang, and S.~Yan, ``Image
  classification with tailored fine-grained dictionaries,'' \emph{IEEE Trans.
  Circuit Syst. Video Technol.}, vol.~28, no.~2, pp. 454--467, 2016.

\bibitem{peng2018object}
Y.~{Peng}, X.~{He}, and J.~{Zhao}, ``Object-part attention model for
  fine-grained image classification,'' \emph{IEEE Trans. Image Process.},
  vol.~27, no.~3, pp. 1487--1500, 2018.

\bibitem{zheng2020learn}
H.~{Zheng}, J.~{Fu}, Z.~{Zha}, J.~{Luo}, and T.~{Mei}, ``Learning rich part
  hierarchies with progressive attention networks for fine-grained image
  recognition,'' \emph{IEEE Trans. Image Process.}, vol.~29, pp. 476--488,
  2020.

\bibitem{lin2018bilinear}
T.~{Lin}, A.~{RoyChowdhury}, and S.~{Maji}, ``Bilinear convolutional neural
  networks for fine-grained visual recognition,'' \emph{IEEE Trans. Pattern
  Anal. Mach. Intell.}, vol.~40, no.~6, pp. 1309--1322, 2018.

\bibitem{xiao2015learning}
T.~Xiao, T.~Xia, Y.~Yang, C.~Huang, and X.~Wang, ``Learning from massive noisy
  labeled data for image classification,'' in \emph{IEEE Conf. Comput. Vis.
  Pattern Recog.}, 2015, pp. 2691--2699.

\bibitem{ZhangBHRV17}
C.~Zhang, S.~Bengio, M.~Hardt, B.~Recht, and O.~Vinyals, ``Understanding deep
  learning requires rethinking generalization,'' in \emph{Int. Conf. Learn.
  Represent.}, 2017.

\bibitem{fgsm}
I.~J. Goodfellow, J.~Shlens, and C.~Szegedy, ``Explaining and harnessing
  adversarial examples,'' in \emph{Int. Conf. Learn. Represent.}, 2015.

\bibitem{pgd}
A.~Kurakin, I.~J. Goodfellow, and S.~Bengio, ``Adversarial machine learning at
  scale,'' in \emph{Int. Conf. Learn. Represent.}, 2017.

\bibitem{Peterson_2019_ICCV}
J.~C. Peterson, R.~M. Battleday, T.~L. Griffiths, and O.~Russakovsky, ``Human
  uncertainty makes classification more robust,'' in \emph{Int. Conf. Comput.
  Vis.}, 2019, pp. 9616--9625.

\bibitem{redmon2016you}
J.~Redmon, S.~Divvala, R.~Girshick, and A.~Farhadi, ``You only look once:
  Unified, real-time object detection,'' in \emph{IEEE Conf. Comput. Vis.
  Pattern Recog.}, 2016, pp. 779--788.

\bibitem{CRM2020Fang}
F.~{Fang}, L.~{Li}, H.~{Zhu}, and J.~{Lim}, ``Combining faster r-cnn and
  model-driven clustering for elongated object detection,'' \emph{IEEE Trans.
  Image Process.}, vol.~29, pp. 2052--2065, 2020.

\bibitem{FPR2019Sun}
F.~{Sun}, T.~{Kong}, W.~{Huang}, C.~{Tan}, B.~{Fang}, and H.~{Liu}, ``Feature
  pyramid reconfiguration with consistent loss for object detection,''
  \emph{IEEE Trans. Image Process.}, vol.~28, no.~10, pp. 5041--5051, 2019.

\bibitem{faster2017Ren}
S.~{Ren}, K.~{He}, R.~{Girshick}, and J.~{Sun}, ``Faster r-cnn: Towards
  real-time object detection with region proposal networks,'' \emph{IEEE Trans.
  Pattern Anal. Mach. Intell.}, vol.~39, no.~6, pp. 1137--1149, 2017.

\bibitem{liu2016ssd}
W.~Liu, D.~Anguelov, D.~Erhan, C.~Szegedy, S.~Reed, C.-Y. Fu, and A.~C. Berg,
  ``Ssd: Single shot multibox detector,'' in \emph{Eur. Conf. Comput. Vis.},
  2016, pp. 21--37.

\bibitem{tian2019fcos}
Z.~Tian, C.~Shen, H.~Chen, and T.~He, ``Fcos: Fully convolutional one-stage
  object detection,'' in \emph{Int. Conf. Comput. Vis.}, 2019, pp. 9627--9636.

\bibitem{everingham2010pascal}
M.~Everingham, L.~Van~Gool, C.~K. Williams, J.~Winn, and A.~Zisserman, ``The
  pascal visual object classes (voc) challenge,'' \emph{Int. J. Comput. Vis.},
  vol.~88, no.~2, pp. 303--338, 2010.

\bibitem{guo2017calibration}
C.~Guo, G.~Pleiss, Y.~Sun, and K.~Q. Weinberger, ``On calibration of modern
  neural networks,'' in \emph{Int. Conf. Mech. Learn.}, 2017, pp. 1321--1330.

\end{thebibliography}
\ifCLASSOPTIONcaptionsoff
  \newpage
\fi

\vfill

\end{document}